%% file: iclr2025_conference.tex
\documentclass{article} 
\usepackage{iclr2025_conference,times}
\usepackage{CJKutf8}
\usepackage{xpinyin}
\usepackage[T1]{fontenc}

\input{math_commands.tex}

\usepackage{hyperref}
\usepackage{url}
\usepackage{graphicx} 
\usepackage{subfigure}
\usepackage{booktabs}
\usepackage{times}
\usepackage{epsfig}
\usepackage{graphicx}
\usepackage{amsmath}
\usepackage{amssymb}
\usepackage{pifont}
\usepackage{booktabs}
\usepackage{arydshln}
\usepackage{color}
\usepackage{colortbl}
\usepackage{subcaption}
\usepackage{multirow}
\usepackage{float}
\usepackage{rotfloat}
\usepackage{capt-of}
\usepackage{longtable}
\usepackage{diagbox}
\usepackage{makecell}
\usepackage{xspace}
\usepackage{wrapfig}
\usepackage{enumitem}
\usepackage{fontawesome}
\usepackage{epigraph}
\usepackage{multirow}
\usepackage{multicol}
\usepackage{rotating}
\usepackage{amssymb}
\usepackage{pifont}
\usepackage{algorithm}
\usepackage{algpseudocode}
\usepackage{colortbl}
\usepackage{tcolorbox}
\usepackage{fbox}

\tcbuselibrary{most}

\def\modelname{TextHawk2\xspace}

\title{\modelname: A Large Vision-Language Model Excels in Bilingual OCR and Grounding with 16x Fewer Tokens}

\author{Ya-Qi Yu, Minghui Liao, Jiwen Zhang, Jihao Wu\\
Huawei Technologies Co., Ltd.\\
\texttt{\{yuyaqi5, liaominghui1\}@huawei.com}
}

\iclrfinalcopy 
\begin{document}

\maketitle

\begin{abstract}
Reading dense text and locating objects within images are fundamental abilities for Large Vision-Language Models~(LVLMs) tasked with advanced jobs. Previous LVLMs, including superior proprietary models like GPT-4o, have struggled to excel in both tasks simultaneously. Moreover, previous LVLMs with fine-grained perception cost thousands of tokens per image, making them resource-intensive. We present \modelname, a bilingual LVLM featuring efficient fine-grained perception and demonstrating cutting-edge performance across general-purpose, OCR, and grounding tasks with 16 times fewer image tokens. Critical improvements include: (1)~Token Compression: Building on the efficient architecture of its predecessor, \modelname significantly reduces the number of tokens per image by 16 times, facilitating training and deployment of the TextHawk series with minimal resources. (2)~Visual Encoder Reinforcement: We enhance the visual encoder through LVLM co-training, unlocking its potential for previously unseen tasks like Chinese OCR and grounding. (3)~Data Diversity: We maintain a comparable scale of 100 million samples while diversifying the sources of pre-training data. We assess \modelname across multiple benchmarks, where it consistently delivers superior performance and outperforms closed-source models of similar scale, such as achieving 78.4\% accuracy on OCRBench, 81.4\% accuracy on ChartQA, 89.6\% ANLS on DocVQA, and 88.1\% accuracy@0.5 on RefCOCOg-test.
\end{abstract}

\begin{figure}[H]
  \centering
  \includegraphics[width=\textwidth]{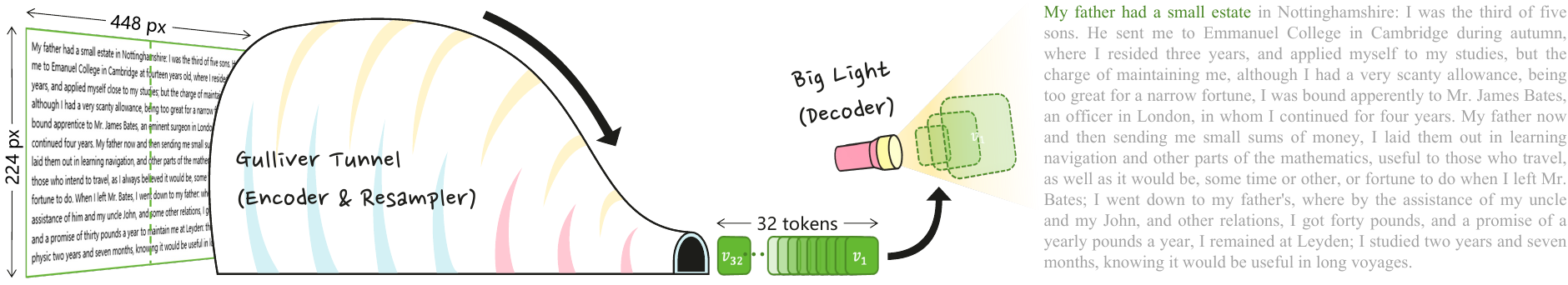}
  \caption{The magic of visual token compression. In this demonstration, \modelname compresses 183 words displayed on a $448 \times 224$ image, where each character measures under 8 pixels, into 32 tokens, allowing for accurate recovery. It’s reminiscent of the futuristic gadgets in \textit{Doraemon} anime.}
  \label{fig:compress}
\end{figure}

\section{Introduction}
\label{sec:intro}
Over the past few years, significant advancements have been made in the realm of Large Language Models~(LLMs)~\citep{llama,glm4,qwen2,deepseek-v2,internlm2}. These breakthroughs have also driven the development of Large Vision-Language Models~(LVLMs)~\citep{blip2,llava,cogvlm,qwen-vl,deepseek-vl,internvl-1.5}. LVLMs effectively combine visual and linguistic modalities, allowing them to understand visual content while leveraging the instruction-following and dialogue capabilities of LLMs. Over the past year, the rapid evolution of LVLMs, incorporating larger foundational LLMs and richer datasets, has significantly improved their ability to perform complex multimodal understanding and reasoning. Consequently, state-of-the-art LVLMs have achieved outstanding results across various Visual Question Answering (VQA) benchmarks. However, to apply LVLMs to real-world scenarios beyond general VQA tasks, there is a need for a more refined perception of visual details, such as Optical Character Recognition (OCR) and object localization~\citep{uihawk}. These advanced capabilities are essential for applications in areas like document intelligence, Graphics User Interface~(GUI) agents, and visual assistance for blind and low-vision users, where accurately interpreting and responding to detailed visual information is critical.

Recent advancements in leading LVLMs have significantly improved their ability to recognize and interpret dense text within images~\citep{llavanext-strong,mplug-docowl-1.5,glm4,ixc2-4khd,internvl-1.5}. GPT-4V, as the initial multimodal version of ChatGPT, demonstrates strong OCR capabilities for English text. However, it struggles with accurately interpreting Chinese text, often leading to hallucinations. This limitation has been significantly relieved in its successor, GPT-4o, which substantially improves its performance in handling non-English text, including Chinese. In the open-source domain, significant strides have also been made with models like InternVL series~\citep{internvl-1.5}. For instance, InternVL 1.2 increases its image resolution from 224 to 448, allowing it to capture finer details. This improvement is complemented by co-training the visual encoder on a mix of image captioning and OCR-specific datasets, boosting the model's ability to recognize text effectively within images. Building on this progress, InternVL 1.5 employes an image cropping strategy that enables the dynamic processing of high-resolution images.

However, text-oriented LVLMs often require processing a large number of tokens when handling high-resolution images, which results in significant computational costs and extensive context usage. This is due to the rapid increase in image tokens, making it crucial to compress them effectively. Despite this need, previous top-performing OCR models have only achieved an image compression ratio of up to 4, which is inadequate for practical applications. This raises the first question: \textit{Can we increase the compression ratio to 16 without losing the ability to perceive fine-grained details and achieve state-of-the-art OCR performance with limited resources?}

Despite the impressive visual understanding and OCR performance shown by leading models like GPT-4o and InternVL 1.5, they still face challenges in achieving basic grounding capabilities. Unlike generalist LVLMs that often employ language-supervised visual encoders, grounding-oriented models~\citep{ground-dino,ferret} typically rely on self-supervised visual encoders like DINOv2~\citep{dinov2}. However, an intriguing finding is that language-supervised visual encoders actually outperform self-supervised ones, particularly on OCR tasks, where the gap is notably wide~\citep{cambrian-1}. Some studies~\citep{ferret,sphinx} have suggested combining multiple visual encoders, like CLIP~\citep{clip} and DINOv2, to improve performance. Nonetheless, while these models aim to improve grounding capabilities, none guarantees strong performance across general multimodal understanding and OCR tasks. Additionally, using dual encoders leads to computational redundancy. This leads to the second question: \textit{Can we train an LVLM with a single visual encoder that excels in general multimodal understanding, OCR, and grounding simultaneously?}

In this study, we delve into the previously mentioned questions, aiming to provide a comprehensive analysis and innovative solutions. Our key contributions are summarized as follows:
\begin{itemize}
  \item We introduce \modelname, a versatile LVLM that accommodates visual inputs of any resolution and demonstrates outstanding performance on fine-grained benchmarks, including OCRBench, ChartQA, DocVQA, InfoVQA, RefCOCO, and others.
  \item We demonstrate that our thoughtfully designed resampler can compress visual tokens by a factor of 16 without compromising fine-grained perception capabilities.
  \item We establish that, through effective data curation and reinforcement of the visual encoder, it is possible to achieve state-of-the-art performance in general multimodal understanding, OCR, and grounding simultaneously with a unified visual encoder.
\end{itemize}

\section{Related Works}
\label{sec:related}

\subsection{Text-Oriented LVLMs}
Text recognition or document understanding is a pivotal feature of LVLMs. Consequently, numerous LVLMs dedicate their efforts not only to general image comprehension but also to text-oriented tasks. Initially, certain methodologies, such as LLaVAR~\citep{llavar} and mPLUG-DocOwl~\citep{mplugdocowl}, enhance image resolution and incorporate text-rich data during the instruction tuning phase. For instance, mPLUG-DocOwl elevates the image resolution to $896 \times 896$ and integrates a diverse array of text-rich data, including documents, tables, webpages, and charts, building upon the mPLUG-Owl~\citep{mplugowl} framework. CogAgent~\citep{cogagent}, on the other hand, employs both low-resolution and high-resolution image encoders to accommodate inputs at a resolution of $1120 \times 1120$. Subsequently, UReader~\citep{ureader} introduces a shape-adaptive cropping module tailored for handling high-resolution images. mPLUG-DocOwl 1.5~\citep{mplug-docowl-1.5} adopts this cropping module and constructs an extensive dataset, DocStruct4M, to further refine its text-oriented capabilities. InternLM-XComposer2-4KHD~\citep{ixc2-4khd} ventures into image resolutions up to 4K HD and beyond, employing a similar cropping strategy. Meanwhile, InternVL 1.5~\citep{internvl-1.5} incorporates OCR data during the pre-training phase, thereby significantly enhancing the models' text recognition capabilities.

TextHawk~\citep{texthawk1} adheres to the shape-adaptive cropping strategy for handeling arbitrary shape images and introduces innovative features such as Scalable Positional Embeddings~(SPEs) and Query Proposal Network~(QPN) to more effectively model sub-images. Additionally, it incorporates a Multi-Level Cross-Attention~(MLCA) mechanism that capitalizes on the hierarchical structure and semantic relationships within the data, thereby significantly enhancing the model's fine-grained visual perception capabilities. \modelname builds upon the architecture of TextHawk, with enhancements made in both the training data and the model's training strategies, aiming to achieve superior performance in text-oriented tasks.

\subsection{Grounding-Oriented LVLMs}
Grounding capabilities are critical for LVLMs to tackle complex reasoning tasks involving specific regions and objects within images. To improve interpretability and enhance user interaction, LVLMs are typically expected to accept and provide positional information in formats such as point coordinates, bounding boxes, or region masks. Shikra~\citep{shikra} approaches this by encoding positions as normalized plain-text coordinates, leveraging the flexibility of natural language. Conversely, models like VisionLLM~\citep{visionllm}, Kosmos-2~\citep{kosmos2}, and Ferret~\citep{ferret} extend LVLM vocabularies by incorporating location tokens that represent normalized and quantized offsets of image dimensions. These models are trained on carefully crafted large-scale visual grounding datasets to support the new tokens. LLaVA-G~\citep{llava-grounding} adopts a different strategy, predicting segmentation masks rather than bounding boxes, using a pre-trained grounding model as its decoder, which necessitate additional alignment training with the LVLM. Meanwhile, GPT4RoI~\citep{gpt4roi} and VolCano~\citep{volcano} enhance fine-grained multimodal understanding by supplementing the model with additional regional features, instead of positional information. In contrast, TextHawk series pioneer the native grounding capability of LVLMs via a detection head combined with efficient representation of bounding boxes.

\subsection{Visual Token Compression}
With the support for higher image resolutions in recent LVLMs, the number of visual tokens has surged, creating a strong demand for efficient compression methods. Solutions like CogAgent and MiniGemini~\citep{minigemini} tackle this by introducing a lightweight visual encoder specifically for high-resolution refinement, without increasing the visual token count. Their method uses low-resolution visual embeddings as queries to retrieve relevant high-resolution cues, either within the LLM or via a resampler. Qwen-VL~\citep{qwen-vl} and LLaVA-UHD~\citep{llavauhd} adopt a different approach by directly compressing visual tokens of each sub-image by factors of 4 and 9, respectively, using a shared perceiver resampler layer. Meanwhile, LLaVA-PruMerge~\citep{llavaprumerge} implements an adaptive strategy, dynamically identifying and retaining the most critical visual tokens, then merging similar ones through clustering. TextMonkey~\citep{textmonkey} also performs visual token compression based on token similarity. MADTP~\citep{madtp} introduces a Dynamic Token Pruning (DTP) module that adjusts visual token compression ratios layer by layer, adapting to varying input complexities. TextHawk stands out as the first LVLM to achieve 16 times token compression through a novel two-step process for resampling and rearrangement, each reducing the token count by a factor of 4. Building on TextHawk’s approach, \modelname achieves the same compression ratio of 16, offering enhanced efficiency in visual token handling.

\section{Architecture}
\label{sec:arch}

\begin{figure}
  \centering
  \includegraphics[width=\textwidth]{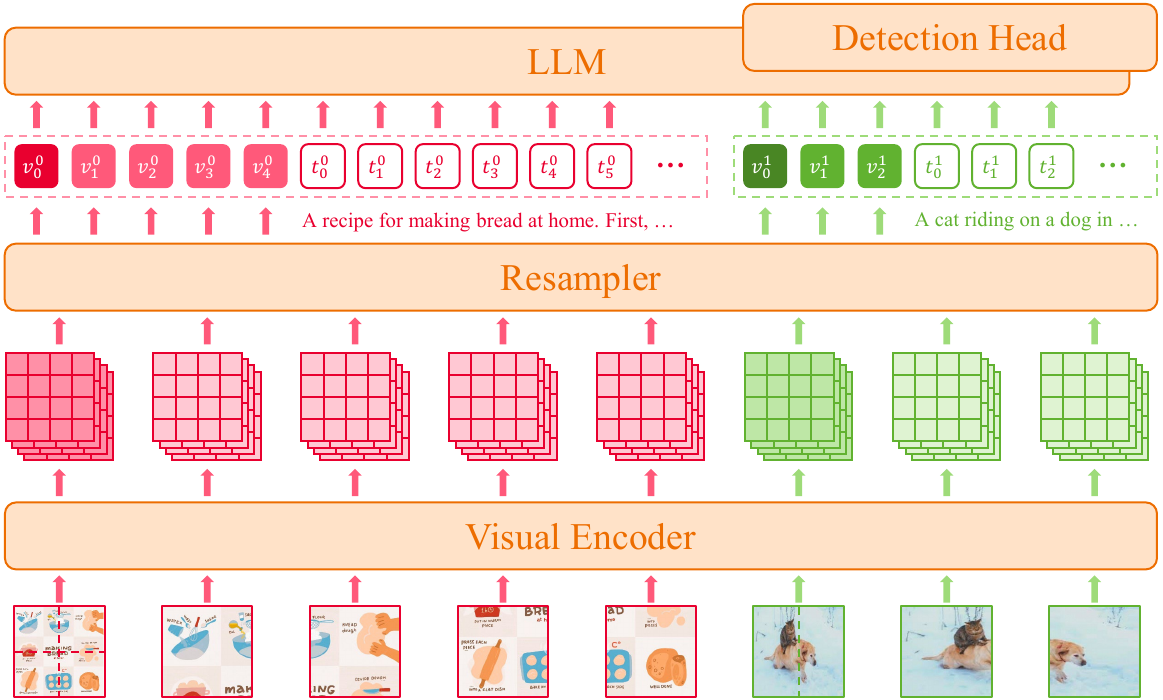}
  \caption{The network architecture and dataflow of \modelname.}
  \label{fig:arch}
\end{figure}

The overall architecture and key components of \modelname continue the design of TextHawk~\citep{texthawk1} family, including a lightweight visual encoder and an LLM, which are bridged by a meticulously designed resampler for modality adaption and token compression. The network architecture and dataflow of \modelname is depicted in Fig.~\ref{fig:arch}.

\subsection{Large Language Model}
\label{ssec:llm}
Recent advances in open-source LLMs have also enhanced the ability of upper-level LVLMs to understand both linguist and visual content. Noteworthy among these developments are models like LLaMA3~\citep{llama3} and Qwen2~\citep{qwen2}. There are two major enhancements in the latest LLMs. Firstly, the integration of Grouped-Query Attention~(GQA) has greatly reduced the memory requirements for key-value cache during deployment. Secondly, these models support longer context lengths, e.g., Qwen2 can handle up to 32,768 tokens during pre-training and extend up to 131,072 tokens during inference. To develop a Chinese-English bilingual LVLM, we utilize Qwen2-7B-Instruct due to its strong capability in processing Chinese data.

\subsubsection{Detection Head}
\label{ssec:det}
\begin{figure}[H]
  \centering
  \includegraphics[width=0.5\linewidth]{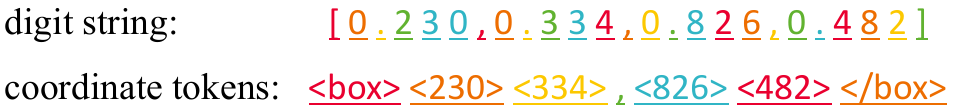}
  \caption{Different representations of coordinates. For clarity, we separate the tokens by different colors and broken underlines.}
  \label{fig:coords}
\end{figure}

To improve training throughput and inference efficiency, our TextHawk series extend the vocabulary of LVLMs with special coordinate tokens. As depicted in Fig.~\ref{fig:coords}, representing a bounding box with a digit string requires 25 tokens—2 trigger marks, 4 floating-point numbers~(each uses 5 tokens), and 3 commas. In contrary, by replacing each floating-point number with a unique coordinate token and retaining the center comma, we significantly reduce the token count to just 7.

To facilitate the training of newly appended coordinate tokens and strengthen grounding capability, TextHawk series introduce an auxiliary training objective. This is achieved by integrating a detection head and $\ell_1$ loss. Specifically, the detection head consists of a 2-layer MLP and a linear projection layer, running parallel to the original output layer of the LLM.

\subsection{Visual Encoder}
\label{ssec:vit}
Following TextHawk, we utilize the lightweight ViT from SigLIP-SO400M~\citep{siglip} as the visual encoder of \modelname, maintaining the original resolution of $224\times224$. The effectiveness of the SigLIP family serving as visual encoders for LVLMs has also been demonstrated in concurrent studies~\citep{bunny,idefics2}. Our work further confirms the feasibility of transferring SigLIP to previously unseen tasks, such as Chinese OCR.

\begin{tcolorbox}[title={Unified Visual Encoder},fonttitle=\bfseries,fontupper=\itshape]
  Language-supervised models such as CLIP~\citep{clip} and SigLIP are not optimized for fine-grained tasks. There was once a trend for LVLMs to use dual or even more visual encoders~\citep{sphinx,mousi}. In the realm of text-oriented LVLMs, some of the previous works~\citep{vary,deepseek-vl} use CLIP-ViT as the low-resolution visual encoder and SAM~\citep{sam} as the high-resolution encoder encoder. As for grounding tasks, previous state-of-the-art methods~\citep{sphinx,ferret2} opt for a combination of CLIP-ViT and DINOv2~\citep{dinov2}.
  \tcblower
  Despite the marginal advantages on academic benchmarks, using dual visual encoders is computationally expensive and lacks flexibility in building generalist models, making it impractical for real-world applications. Hence, we opt for a unified visual encoder and enhance it through feature merging~(Section~\ref{sssec:mlca}) and LVLM co-training to maximize its potential.
\end{tcolorbox}

\subsubsection{Dynamic High-Resolution}
\label{sssec:resolution}
Following UReader~\citep{ureader}, the TextHawk family enhances a fixed-resolution ViT through a dynamic cropping strategy. This approach, widely adopted in recent research, effectively processes images with varying aspect ratios and resolutions. For further details, readers may refer to TextHawk~\citep{texthawk1}. Our findings indicate that input resolution significantly impacts the accuracy in fine-grained tasks, particularly OCR tasks like image-to-markdown. For example, low-resolution images of math formulas sometimes contain small and blurry characters, leading to hallucinations. During pre-training, \modelname is configured to allow a maximum area of 36 sub-images and a maximum side length of 12 sub-images per row or column. This setup yields a maximum of 1.8 million pixels and a long edge length of 2688 pixels. Unlike TextHawk, \modelname expands the maximum area from 36 to 72 during supervised fine-tuning, enabling higher-resolution image inputs. To clarify, the maximum area value is relevant only for high-resolution images that surpass the specified limit. For example, an input image with dimensions of $896 \times 672$ will be split into $4 \times 3$ sub-images rather than $8 \times 6$ sub-images, thereby avoiding unnecessary computational costs.

It is important to note that increasing the maximum area can enhance high-resolution performance but introduces two notable side effects. Firstly, it demands a significantly larger amount of memory to store a large batch of ViT activations, which can strain system resources. Secondly, it leads to an unbalanced computational load across different samples due to the varying number of sub-images, which creates inefficiencies. These factors can severely impact training throughput, particularly when using pipeline parallelism with limited resources. Therefore, it is crucial to impose a proper limit on the maximum area to mitigate these issues while balancing high-resolution performance.

\subsection{Resampler}
\label{ssec:resampler}
The resampler plays a critical role in bridging different modalities as well as compressing tokens between the visual encoder and the LLM. For the reader's convenience, we briefly revisit several key improvements of the TextHawk resampler.

\subsubsection{Scalable Positional Embeddings}
\label{sssec:spe}
Scalable Positional Embeddings (SPEs)~\citep{texthawk1} present an innovative extension of factorized positional embeddings~(decomposing row and column), making them applicable to arbitrary input shapes. To ensure a fair comparison, we also modify Absolute Positional Embeddings (APEs) to accommodate dynamic shapes by slicing sections of the APEs during both training and inference phases. Due to their adaptability and training efficiency, SPEs achieve superior performance over APEs while utilizing fewer parameters. Furthermore, SPEs exhibit outstanding extrapolation capabilities to unseen input resolutions, resulting in impressive zero-shot performance.

The concept of SPEs arises from the observation that positional embeddings, in practice, tend to distribute themselves around the surface of a hypersphere. This insight leads us to consider Spherical Linear Interpolation (Slerp) as a potential alternative to traditional interpolation methods, such as nearest-neighbor or linear interpolation. However, our initial attempts to directly apply Slerp to pre-trained APEs prove to be ineffective. We believe this ineffectiveness stems from the incomplete assumption that these embeddings are perfectly distributed on a hypersphere. To address this issue, we introduce a normalization and scaling process for the embeddings prior to interpolation, ensuring they conform to the requirements of Slerp. Moreover, given that different parts of the positional embeddings are used independently by each attention head, we apply normalization and scaling operations on a per-head basis, allowing for more precise interpolation aligned with the needs of each attention mechanism. The pseudocode of SPEs is shown in Algorithm~\ref{alg:spe}.

\begin{algorithm}[H]
  \caption{Scalable Positional Embeddings}
  \label{alg:spe}
  \begin{algorithmic}[1]
    \Require start embeddings $\ve_0\in\sR^d$, end embeddings $\ve_1\in\sR^d$, interpolation position $t\in[0,1]$
    \Ensure interpolated positional embeddings $\ve(t)$
    \State \textbf{initialization:} {$s\leftarrow\sqrt{d}$}  \Comment{scaling factor}
    \For {$i\in\{0,1\}$}
      \State {$\ve_i\leftarrow\frac{\ve_i}{\|\ve_i\|}$}  \Comment{normalization}
      \State {$\ve_i\leftarrow s\cdot\ve_i$}  \Comment{scaling}
    \EndFor
    \State {$\theta\leftarrow\arccos\frac{\ve_0\ve_1}{\|\ve_0\|\|\ve_1\|}$}
    \State {$\ve(t)\leftarrow\frac{\sin(\theta-t\theta)}{\sin\theta}\ve_0+\frac{\sin(t\theta)}{\sin\theta}\ve_1$}
  \end{algorithmic} 
\end{algorithm}

\subsubsection{Query Proposal Network}
\label{sssec:qpn}
To enhance convergence and improve grounding performance, the TextHawk family incorporates the Query Proposal Network~(QPN)~\citep{texthawk1} to dynamically generate resampling query tokens. Attention-based adapters, such as Quering Former~(Q-Former)~\citep{blip2} and perceiver resampler~\citep{flamingo}, show promise in token compression but are challenging to train. On the other hand, MLP-based adapters, while simpler, often outperform attention-based adapters when training data is limited. We attribute the difficulty of training attention-based adapters to the fixed query tokens used in previous approaches. This observation led us to merge the strengths of both methods. Specifically, QPN utilizes a lightweight MLP-Pool-Dense architecture to efficiently transform features from the visual encoder into queries. It also offers greater adaptability by allowing a variable number of unique queries for images with different resolutions. In the QPN, we apply a $2\times2$ max pooling, achieving a compression ratio of 4 during the resampling stage.

\subsubsection{Resampling and Rearrangement}
\label{sssec:resa}
TextHawk introduces a two-stage token compression strategy called ReSampling and ReArrangement~(ReSA)~\citep{texthawk1}, designed to minimize information loss and preserve critical information from visual inputs. In the first stage, resampling, a smaller set of highly informative tokens is selectively extracted from the visual encoder outputs. This is achieved through a cross-attention mechanism where query tokens, generated by the QPN, guide the selection process. For \modelname, these tokens are progressively refined in 4 bidirectional decoder layers. In the second stage, rearrangement, visual tokens are flattened following the image scanning order and then grouped into sets of four. Instead of arranging tokens based on the sequence of sub-images, we preserve the original line-by-line scanning order of the entire image. The former approach leads to an input order that diverges significantly from the natural reading order of text, which is typically from left to right and top to bottom, thereby impairing the document understanding capabilities of LVLMs. In contrast, our approach arranges visual tokens from sub-images in the same row in an interleaved pattern. Furthermore, our token concatenation strategy aligns with this approach by combining four adjacent tokens within a $1\times4$ window along the same row.

\begin{tcolorbox}[title={16 Times Token Compression},fonttitle=\bfseries]
  By combining the $2 \times 2$ subsampling window from the resampling stage with the $1 \times 4$ subsampling window from the rearrangement stage, we achieve a total compression ratio of 16 within a $2 \times 8$ subsampling window. This window shape may be particularly suited for text-oriented LVLMs, and efforts to explore different window shapes are discussed in a concurrent work~\citep{mplug-docowl-1.5}.
\end{tcolorbox}

\subsubsection{Multi-Level Cross-Attention}
\label{sssec:mlca}
To address the limitations of language-supervised visual encoders on fine-grained tasks, TextHawk proposes a feature merging approach called Multi-Level Cross-Attention~(MLCA)~\citep{texthawk1}. The MLCA mechanism is designed to enhance feature extraction by allowing the resampler to efficiently aggregate information from multiple layers of a visual encoder. This is achieved through a predefined routing table that determines which features are to be extracted and merged at each resampler layer. One of the key findings is that a deep-to-shallow feature extraction strategy yields superior results for grounding tasks while preserving the overall performance of general visual understanding. Notably, MLCA accomplishes this without incurring any additional computational costs, making it both effective and efficient. In practical terms, the implementation of MLCA in TextHawk involves utilizing four distinct stages of the visual encoder. Features are extracted specifically from the 14th, 18th, 22nd, and 26th layers of the encoder.

\section{Data}
\label{sec:data}
\modelname employs a one-pass pre-training approach, differing from the two-stage pre-training paradigm commonly used in prior works on LVLMs. These models undergo an initial stage where different modalities are aligned using fixed, low-resolution image-caption pairs. This is followed by a second stage of continual training on mixed-resolution image-text data from diverse sources, such as OCR and grounding datasets. In contrast, \modelname skips the initial alignment stage and instead focuses on training on more detailed image captions from the beginning.

\subsection{Pre-Training}
\label{ssec:pretrain_data}

\begin{figure}[t]
  \centering
  \includegraphics[width=0.68\linewidth]{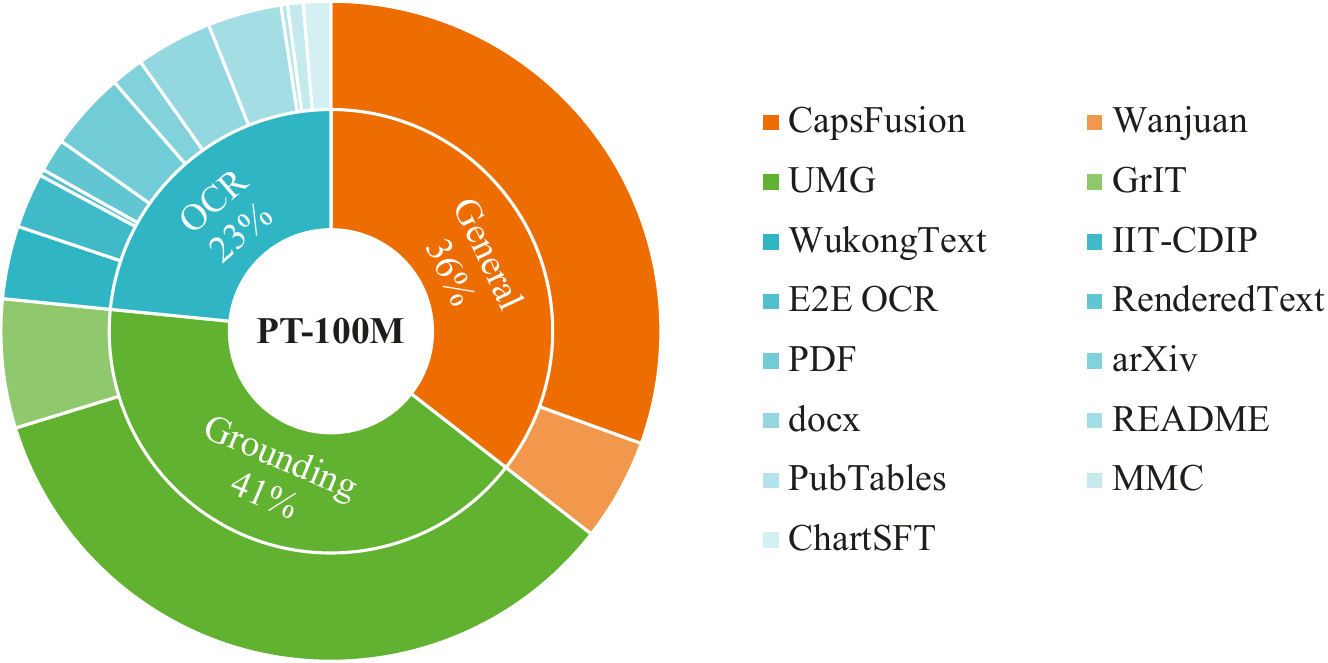}
  \caption{The 100M pre-training data mixture of {\modelname}.}
  \label{fig:data}
\end{figure}

The 100M pre-training data are collected from diverse sources and carefully curated to enhance the OCR and grounding capabilities. The sampling ratios for various datasets are shown in Fig.~\ref{fig:data}.

\subsubsection{Conception}
To improve alignment, we utilize data from CapsFusion~\citep{capsfusion}, a framework designed for re-captioning web-crawled data. Within our previous work, the largest conceptual caption dataset, LAION-400M~\citep{laion400m}, is automatically gathered from the web. This approach can result in captions containing irrelevant descriptions or lacking essential details, causing hallucinations and misalignments. CapsFusion addresses these issues by employing LVLM to generate captions that directly reflect the image content. These generated captions are then integrated with the web-sourced captions using a caption fuser, avoiding knowledge loss.

\subsubsection{Interleaved}
Previous works have demonstrated that interleaved image-text data is beneficial for improving the multimodal in-context learning capability of LVLMs~\citep{ixc,idefics2}. \modelname makes up for the lack of large-scale interleaved data by leveraging the image-text dataset from the Wanjuan1.0 data collection~\citep{wanjuan}. This part comprises bilingual interleaved data sourced from Wikipedia and news outlets.

\subsubsection{Grounding}
We primarily utilize GrIT-20M~\citep{kosmos2}, a synthetic caption dataset with additional location labels for major visual elements, to enhance the grounding capability of \modelname. Additionally, we incorporate referring and grounding data from UMG-41M~\citep{umg}. These data are curated from various public image-caption datasets, including CC3M~\citep{cc3m}, CC12M~\citep{cc12m}, SBU~\citep{sbu}, Flickr~\citep{flickr}, VG~\citep{vg}, YFCC-15M~\citep{yfcc}, and ImageNet-21K~\citep{in21k}, by jointly applying an object detector and a regional captioner. Specifically, each region is randomly assigned to either a referring task or a grounding task. In the referring task, we provide the model with a bounding box to generate a caption for that specific region, while in the grounding task, we reverse this by using the caption to predict the corresponding bounding box. We also include approximately 1/8 of the captions in Chinese, which are generated using an English-to-Chinese translation API.

\subsubsection{OCR}
To gather extensive OCR pre-training data, we employ a commercial OCR engine to transcribe text from images. This includes Chinese text from the Wukong~\citep{wukong} dataset and English text from the IIT-CDIP~\citep{iit_cdip} dataset. We also use PDFPlumber to extract text lines from Common Crawl PDFs. To improving English handwriting recognition, we incorporate RenderedText~\citep{renderedtext}. Additional end-to-end OCR datasets, including ArT~\citep{art}, COCO-Text~\citep{cocotext}, CTSU~\citep{ctsu}, CTW~\citep{ctw}, IC15~\citep{ic15}, LSVT~\citep{lsvt}, MLT~\citep{mlt}, MTWI~\citep{mtwi}, RCTW-17~\citep{rctw17}, ReCTS~\citep{rects}, and SCUT-HCCDoc~\citep{hccdoc} are also integrated into our training data.

\subsubsection{Markdown}
Building upon the markup-based data pipeline introduced in Kosmos-2.5~\citep{kosmos2_5}, we expand our dataset by gathering more image-to-markdown pairs to enhance OCR and layout understanding capabilities. We specifically source \LaTeX~documents from arXiv, README files from GitHub, and DOCX files from Common Crawl. These files are then converted into images and subsequently translated into markdown format.

\subsubsection{Table \& Chart}
Alongside the previously mentioned markdown data, we also collect data to enhance the ability to interpret tables and charts. For tables, we use the PubTables-1M~\citep{pubtables} dataset, including both its original English version and a translated Chinese version, to gather table recognition data. For charts, we employ chart-to-table conversion and chart-based QA data from existing datasets, including MMC~\citep{mmc} and ChartSFT~\citep{chartast}.

\subsection{Supervised Fine-Tuning}
\label{ssec:sft_data}
\modelname enhances the mixture of TextHawk instruction data by incorporating several newly added datasets. First, it replaces all text-only data with two high-quality data collections: OpenHermes2.5~\citep{openhermes2_5} and COIG-CQIA~\citep{coig_cqia}. Next, it adds a variety of other datasets, including ShareGPT-4o~\citep{internvl-1.5}, LVIS-Instruct4V~\citep{lvis_instruct4v}, LAION-GPT4V, LLaVAR~\citep{llavar}, Cauldron~\citep{idefics2}, KVQA~\citep{kvqa}, ViQuAE~\citep{viquae}, Geo170K~\citep{g-llava}, HME100K~\citep{hme100k}, UniMER-1M~\citep{unimer}, FUNSD~\citep{funsd}, XFUND~\citep{xfund}, SROIE~\citep{sroie}, POIE~\citep{poie}, ST-VQA~\citep{stvqa}, and EST-VQA~\citep{estvqa}. Finally, it randomly samples 80K, 60K, 20K, 300K, and 80K pre-training data samples from the aforementioned OCR, Markdown, Table, and Chart categories.

\section{Implementation Details}
\label{sec:impl}

\subsection{Infrastructure}
\label{ssec:infra}
\modelname is trained on Huawei Cloud, utilizing the elastic cloud computing and file system.

\begin{figure}[t]
  \centering
  \includegraphics[width=\linewidth]{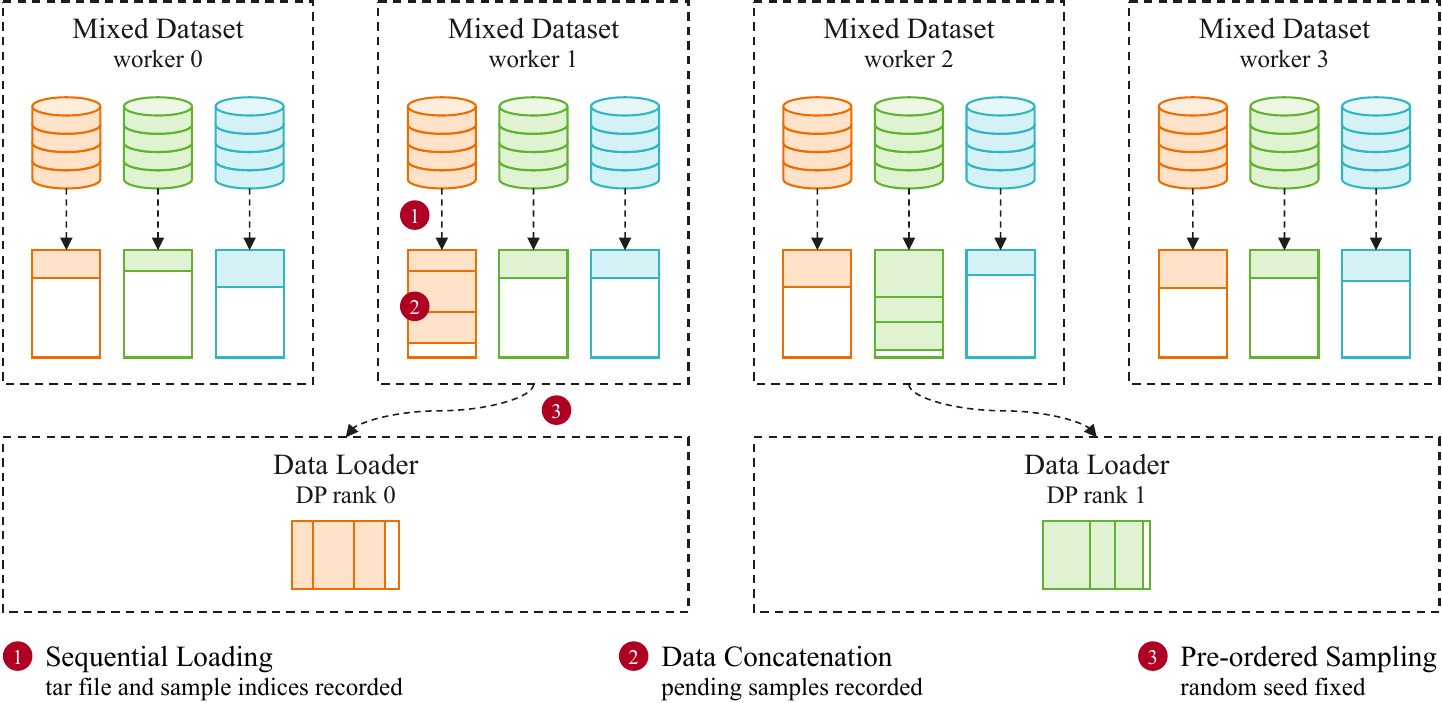}
  \caption{Data pipeline with sequential loading and failure recovery mechanism.}
  \label{fig:data_pipeline}
\end{figure}

\subsubsection{Storage}
\label{sssec:storage}
For storing large-scale multimodal data, we use Huawei Cloud's Parallel File System~(PFS). PFS is a high-performance file system built on Object Storage Service~(OBS), offering millisecond-level access latency, TB/s-level bandwidth, and millions of IO/s, enabling fast handling of High-Performance Computing~(HPC) workloads.

To accelerate data preparation during training, we introduce a sequential loading method that avoids the inefficiency of random access to numerous small files. Our dataset implementation is built on WebDataset~\citep{webdataset}, a high-performance IO system that uses tar files and provides Python APIs. It supports reading files from local storage as well as files from OBS. By using the WebDataset format, we create a fully sequential IO pipeline optimized for handling large-scale data. Specifically, we divide the samples into equal-sized chunks, store them in tar files, and distribute the chunks across different data workers. Each data worker then loads all the samples from its assigned chunks sequentially, with no overlap in the data between workers. This approach is crucial for maximizing IO throughput and efficiently leveraging cloud storage during training.

Additionally, we implement a failure recovery mechanism that ensures training can accurately resume from any checkpoint. While the native WebDataset offers fully indexed access to datasets, it introduces significant storage and IO overhead and is not suitable for training. Hence, we introduce another efficient failure recovery mechanism with three main components. First, we shuffle each dataset before training, removing the need for large cache pools during dynamic shuffling. This approach also allows for independent shuffling of any newly added datasets. Second, we log both the tar file and sample indices, enabling quick recovery by directly jumping to the specific tar file and skipping previously processed samples. Finally, we back up pending samples that dataset workers have processed but not yet sent to the data loaders, as well as the state of random generators to guarantee accurate recovery. The overall data pipeline is shown in Fig.~\ref{fig:data_pipeline}.

\subsubsection{Accelerator}
\label{sssec:npu}
We support both NVIDIA GPU and Ascend NPU platforms. To accelerate attention computation, we adopt memory-efficient attention~\citep{xformers} for NVIDIA V100, and NPU fusion attention~\citep{npu_fusion_attention} for Ascend 910B. \modelname is trained on 16 nodes with 128 cards.

\subsection{Efficiency}
\label{ssec:efficiency}
To enhance the training efficiency of LVLMs on devices with limited memory, we present two major improvements, including 4D parallelism and data packing.

\subsubsection{Parallelism}
\label{sssec:parallel}
We employ a combination of four types of parallelisms, including Data Parallelism (DP)~\citep{ddp}, Tensor Parallelism (TP)~\citep{megatron1}, Sequence Parallelism (SP)~\citep{megatron3}, and Pipeline Parallelism~(PP)~\citep{gpipe,pipedream,megatron2}. DP is the most prevalent technique for distributed training, allowing large data batches to be divided across various devices. We utilize DP alongside the DeepSpeed Zero Redundancy Optimizer~(ZeRO)~\citep{zero} to enhance memory efficiency. TP reduces memory usage by distributing model weights and partial activations across multiple devices, while SP further alleviates memory demands by handling activations that TP cannot manage. However, TP introduces significant communication overhead, requiring all devices to reside on the same node with high-speed connections. By default, we set \(\text{TP} = \text{SP} = 1\), using a maximum of \(\text{TP} = \text{SP} = 4\) only when needed. In contrast to LLM training, PP operates differently in LVLM training due to the heterogeneous characteristics of LVLMs. Here are the challenges:
\begin{itemize}
  \item \textbf{Computational Imbalance}: Distributing model layers evenly across multiple pipeline stages is crucial for load balancing. However, achieving this balance with LVLMs is more complex than with LLMs. The challenge arises from the requirement to place the visual encoder and resampler before the first LLM layer, complicating the even distribution of these components across pipeline stages.
  \item \textbf{Memory Imbalance}: The initial pipeline stages have to store activations from the warm-up micro-batches. The size of these activations is proportional to the number of pipeline stages. As PP increases, the memory required to store activations in both the visual encoder and resampler also increases, which might lead to memory overload.
\end{itemize}

For PP, computational imbalance results in increased idle time (aka bubble size), which should be avoided. To address this and improve communication efficiency, we integrate the entire visual encoder and resampler into the initial pipeline stage. To avoid memory overload, we restrict PP, setting it to $\text{PP}=1$ during LoRA training and $\text{PP}=2$ during full-parameters training. To tackle computational imbalance, we divide the LLM layers into unequal segments, ensuring that the first pipeline stage contains the visual encoder, resampler, and fewer LLM layers.

\subsubsection{Packing}
\label{sssec:pack}
To achieve optimal performance, it is crucial to balance the model components with the data stream. However, LVLMs with variable resolution inputs and variable length outputs inevitably involve imbalances. To address this, we set a fixed context length and pack multiple samples to reduce padding. We also restrict the number of packed images to avoid overloading the visual encoder. Specifically, we use a context length of 4096 and a maximum of 108 image tiles, including thumbnails and sub-images. Additionally, we apply masking to ensure that the samples remain mutually invisible.

\subsection{Hyperparameters}
\label{ssec:hparam}
During the pre-training phase, our focus is on training the newly initialized resampler and updating both the ViT and LLM using LoRA. Specifically, LoRA modules are applied to the query and value projection layers, with ranks of 16 for ViT and 128 for LLM. In the supervised fine-tuning stage, we unfreeze all parameters, allowing the entire model to be trained end-to-end. Our preliminary investigation of different training strategies has shown that training an LVLM for Chinese OCR with a frozen ViT is possible. However, unfreezing ViT during both pre-training and supervised fine-tuning significantly enhances performance, making it essential for achieving state-of-the-art results in OCR tasks. To further improve OCR robustness, we introduce manual perturbations by randomly resizing images from text-oriented datasets within a small range.

During the pre-training phase, we utilize a global batch size of 384, where each data point is a collection of multiple packed samples. The training process spans 45,000 steps. The learning rate initiates at 0 and linearly warms up to $2 \times 10^{-4}$ within the initial 3\% of the steps. Beyond this point, it follows a cosine decay schedule, tapering down to 0. We also incorporate a "late warm-up" strategy for both ViT and the LLM. During the first half of the warm-up phase, the parameters of these modules remain fixed. Concurrently, only the parameters of the resampler are updated, which serves to offset the lack of a dedicated pre-training phase for the resampler alone.
For the supervised fine-tuning stage, the global batch size is set to 256, and the model undergoes training for two epochs. The learning rate schedule is akin to the pre-training phase, albeit with distinct peak values: $5 \times 10^{-5}$ for both the ViT and the resampler, and $2 \times 10^{-5}$ for the LLM.

In configuring the AdamW optimizer for stable training, we set $\beta_1$ to 0.9 and $\beta_2$ to 0.95. Additionally, a weight decay of 0.05 is applied to enhance model generalization.

\section{Experiments}
\label{sec:expr}

\begin{table}[t]
  \small
  \centering
  \caption{Performance comparison on text-oriented tasks.} 
  \label{tab:ocr}
    \setlength{\tabcolsep}{0.8mm}\begin{tabular}{lccccccc}
    \toprule
    \textbf{Model} & OCRBench & ChartQA & DocVQA & InfoVQA & TabFact & WTQ & TextVQA \\
    \hline
    \rowcolor{lightgray!50}\multicolumn{8}{l}{\textcolor{gray}{Low compression ratio}}\\
    Qwen-VL-Chat~\citep{qwen-vl}             & 50.6 & 66.3 & 62.6 & 28.3 & -    & -    & 61.5 \\
    UReader~\citep{ureader}                  & -    & 59.3 & 65.4 & 42.2 & 67.6 & 29.4 & 61.5 \\
    Monkey~\citep{monkey}                    & 51.4 & 65.1 & 66.5 & 36.1 & -    & 25.3 & 67.6 \\
    CogAgent~\citep{cogagent}                & 59.0 & 68.4 & 81.6 & 44.5 & -    & -    & 76.1 \\
    DocOwl-1.5-Chat~\citep{mplug-docowl-1.5} & 59.9 & 70.2 & 82.2 & 50.7 & \textbf{80.2} & 40.6 & 68.6 \\
    MiniCPM-V-2.5~\citep{minicpm-v-2.5}      & 72.5 & 72.1 & 84.8 & 50.8 & -    & -    & 76.6 \\
    GLM-4v-9B~\citep{glm4}                   & 78.6 & 30.1 & 76.5 & 53.1 & -    & -    & 83.0 \\
    InternVL2-7B~\citep{internvl2}           & 79.4 & 83.3 & 91.6 & 74.8 & -    & -    & 77.4 \\
    Qwen2-VL-7B~\citep{qwen2-vl}             & \textbf{84.5} & \textbf{83.0} & \textbf{94.5} & \textbf{76.5} & -    & -    & \textbf{84.3} \\
    \rowcolor{lightgray!50}\multicolumn{8}{l}{\textcolor{gray}{High compression ratio}}\\
    TextMonkey~\citep{textmonkey}            & 56.1 & 66.9 & 73.0 & 28.6 & -    & 31.9 & 64.3 \\
    TextHawk~\citep{texthawk1}               & -    & 66.6 & 76.4 & 50.6 & 71.1 & 34.7 & -    \\
    HRVDA~\citep{hrvda}                      & -    & 67.6 & 72.1 & 43.5 & 72.3 & 31.2 & 73.3 \\
    DocKylin~\citep{dockylin}                & -    & 66.8 & 77.3 & 46.6 & -    & 32.4 & -    \\
    DocOwl-2~\citep{mplug-docowl-2}          & -    & 70.0 & 80.7 & 46.4 & \textbf{78.2} & 36.5 & 66.7 \\
    MM1.5~\citep{mm1.5}                      & 63.5 & 78.6 & 88.1 & 59.5 & 75.9 & 46.0 & \textbf{76.8} \\
    \textbf{TextHawk2}                       & \textbf{78.4} & \textbf{81.4} & \textbf{89.6} & \textbf{67.8} & 78.1 & \textbf{46.2} & 75.1 \\
    \bottomrule
  \end{tabular}
\end{table}

\subsection{OCR Benchmark}
\label{ssec:ocr_bench}
We explore the native OCR capabilities of \modelname across various text-oriented tasks, including nature scene text recognition, document information retrieval, chart comprehension, and table fact-checking. The benchmarks utilized are OCRBench~\citep{ocrbench}, ChartQA~\citep{chartqa}, DocVQA~\citep{docvqa}, InfoVQA~\citep{infovqa}, TabFact~\citep{tabfact}, WTQ~\citep{wtq}, and TextVQA~\citep{textvqa}. Our model is compared against multiple baseline LVLMs with different compression ratios, as illustrated in Table~\ref{tab:benchmark}. Notably, \modelname consistently outperforms other baseline LVLMs with high compression ratios by a significant margin. Among the models evaluated, MM1.5~\citep{mm1.5} comes closest in performance, yet \modelname exceeds it by 14.9\%, 2.8\%, 1.5\%, 8.3\%, 2.2\%, and 0.2\% on OCRBench, ChartQA, DocVQA, InfoVQA, TabFact, and WTQ, respectively. When compared to baseline LVLMs with low compression ratios, \modelname surpasses GLM-4v-9B~\citep{glm4}, MiniCPM-V-2.5~\citep{minicpm-v-2.5}, and other previous models. Although it falls short relative to InternVL2-8B~\citep{internvl2}, the performance gap is small. The notable exception is Qwen2-VL-7B~\citep{qwen2-vl}, a member of the highly effective open-source Qwen2-VL series. Qwen2-VL-7B outperforms other leading LVLMs significantly. We attribute this advantage to its native resolution ViT and its full parameter training approach, which we plan to investigate further in future work. In summary, the results of \modelname demonstrate an definite answer to our first question: It is feasible to achieve cutting-edge OCR performance with a visual token compression ratio of 16, where the keys are visual encoder reinforcement and effective data curation.

\begin{table}[t]
  \small
  \centering
  \caption{Performance comparison (Acc@0.5) on referring expression comprehension tasks.} 
  \label{tab:grounding}
    \setlength{\tabcolsep}{2mm}\begin{tabular}{lcccccccc}
    \toprule
    \multirow{2}{*}{\textbf{Model}} & \multicolumn{3}{c}{\textbf{RefCOCO}} & \multicolumn{3}{c}{\textbf{RefCOCO+}} & \multicolumn{2}{c}{\textbf{RefCOCOg}} \\
    \cmidrule(lr){2-4}\cmidrule(lr){5-7}\cmidrule(lr){8-9}
     & val & test-A & test-B & val & test-A & test-B & val & test \\
    \hline
    \rowcolor{lightgray!50}\multicolumn{9}{l}{\textcolor{gray}{Specialist}}\\
    G-DINO-L~\citep{ground-dino}                  & 90.6 & 93.2 & 88.2 & 82.8 & 89.0 & 75.9 & 86.1 & 87.0 \\
    UNINEXT-H~\citep{uninext}                     & 92.6 & 94.3 & \textbf{91.5} & 85.2 & 89.6 & 79.8 & 88.7 & 89.4 \\
    ONE-PEACE~\citep{one-peace}                   & 92.6 & 94.2 & 89.3 & \textbf{88.8} & 92.2 & 83.2 & 89.2 & 89.3 \\
    \rowcolor{lightgray!50}\multicolumn{9}{l}{\textcolor{gray}{Grounding-oriented}}\\
    OFA-L~\citep{ofa}                             & 80.0 & 83.7 & 76.4 & 68.3 & 76.0 & 61.8 & 67.6 & 67.6 \\
    Shikra~\citep{shikra}                         & 87.0 & 90.6 & 80.2 & 81.6 & 87.4 & 72.1 & 82.3 & 82.2 \\
    Ferret-7B~\citep{ferret}                      & 87.5 & 91.4 & 82.5 & 80.8 & 87.4 & 73.1 & 83.9 & 84.8 \\
    Ferret-v2-7B~\citep{ferret2}                  & \textbf{92.8} & 94.7 & 88.7 & 87.4 & 92.8 & 79.3 & 89.4 & 89.3 \\
    CogVLM$_\text{Grounding}$~\citep{cogvlm}      & \textbf{92.8} & \textbf{94.8} & \textbf{89.0} & \textbf{88.7} & \textbf{92.9} & \textbf{83.4} & \textbf{89.8} & \textbf{90.8} \\
    \rowcolor{lightgray!50}\multicolumn{9}{l}{\textcolor{gray}{Generalist}}\\
    Qwen-VL-Chat~\citep{qwen-vl}                  & 88.6 & 92.3 & 84.5 & 82.8 & 88.6 & 76.8 & 86.0 & 86.3 \\
    TextHawk~\citep{texthawk1}                    & 87.3 & 90.9 & 83.3 & -    & -    & -    & -    & -    \\
    InternVL2-8B~\citep{internvl2}                & 87.1 & 91.1 & 80.7 & 79.8 & 87.9 & 71.4 & 82.7 & 82.7 \\
    MM1.5~\citep{mm1.5}                           & -    & 92.5 & 86.7 & -    & 88.7 & 77.8 & -    & 87.1 \\
    Qwen2-VL-7B~\citep{qwen2-vl}                  & 91.7 & \textbf{93.6} & 87.3 & 85.8 & \textbf{90.5} & 79.5 & 87.3 & 87.8 \\
    \textbf{TextHawk2}                            & \textbf{91.9} & 93.0 & \textbf{87.6} & \textbf{86.2} & 90.0 & \textbf{80.4} & \textbf{88.2} & \textbf{88.1} \\
    \bottomrule
  \end{tabular}
\end{table}

\subsection{Grounding Benchmark}
\label{ssec:rec_bench}
Following previous works~\citep{shikra,qwen-vl,ferret}, we investigate the grounding capabilities of \modelname on three Referring Expression Comprehension~(REC) tasks: RefCOCO, RefCOCO+, and RefCOCOg~\cite{refcoco,refcocog}. As presented in Table~\ref{tab:grounding}, we compare \modelname with both generalist and grounding-oriented models, as well as specialist models. Remarkably, \modelname surpasses all current state-of-the-art generalist LVLMs, including the highly effective Qwen2-VL-7B~\citep{qwen2-vl} and InternVL2-8B~\citep{internvl2}. The performance of \modelname on REC tasks is comparable to that of grounding-oriented LVLMs, some of which employ specialized grounding-oriented visual encoders like DINOv2. Combined with the findings in Table~\ref{tab:benchmark}, this confirms the feasibility of training an LVLM using a unified visual encoder that excels across general multimodal understanding, OCR, and grounding tasks, apparently addressing our second question.

\begin{figure}[t]
  \centering
  \subfigure[]{
    \fbox{\includegraphics[height=1250\in]{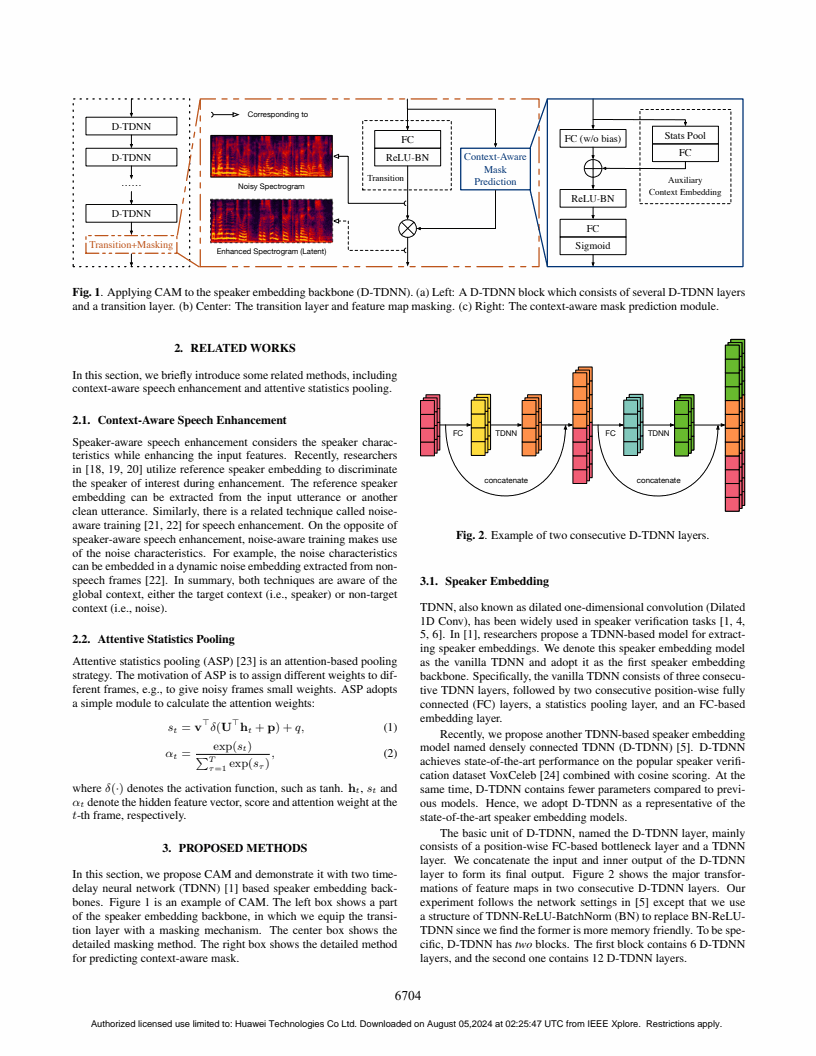}}
    \label{fig:en_img}
  }
  \subfigure[]{
    \fbox{\includegraphics[height=1250\in]{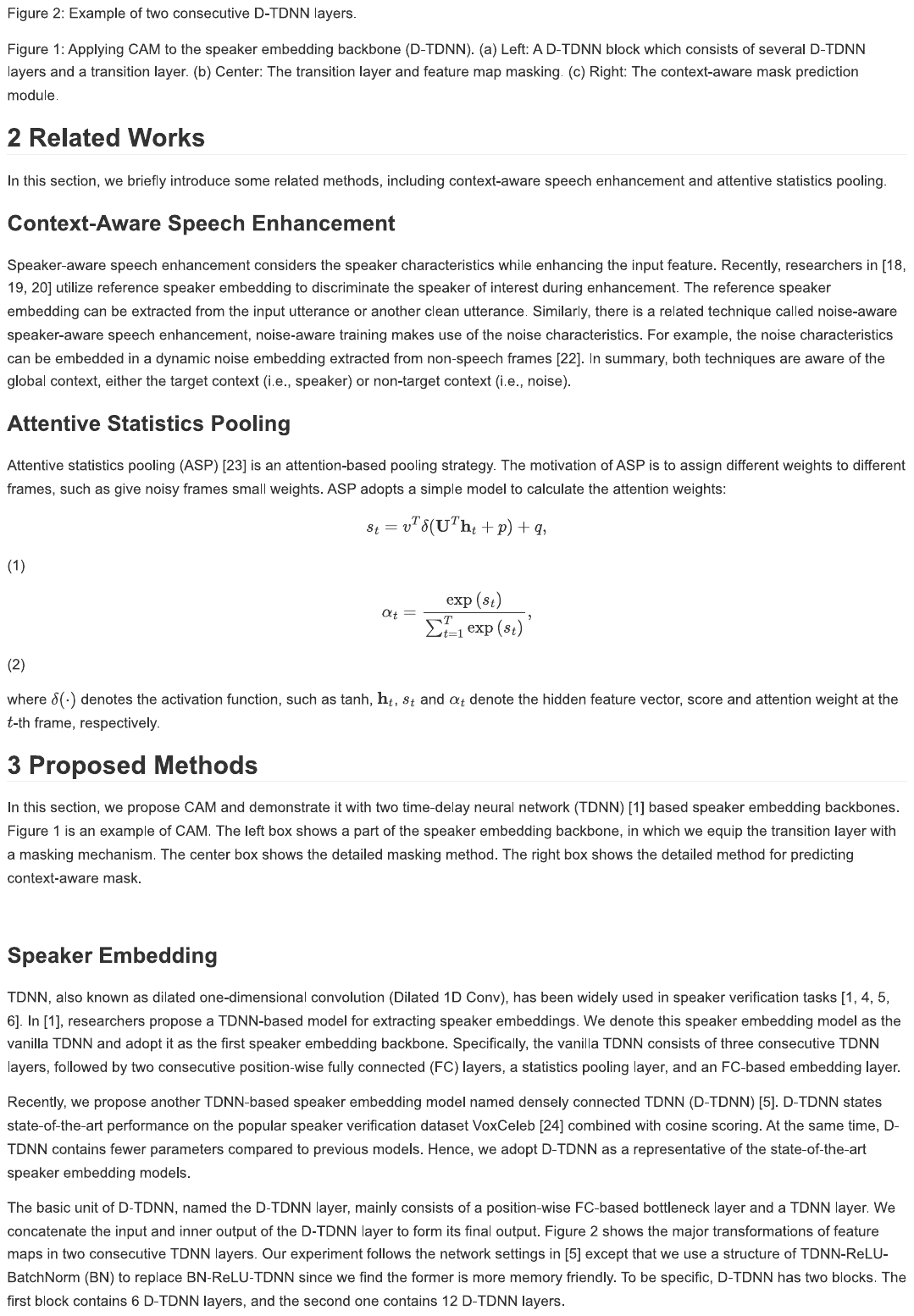}}
    \label{fig:en_md}
  }
  \subfigure[]{
    \fbox{\includegraphics[height=1100\in]{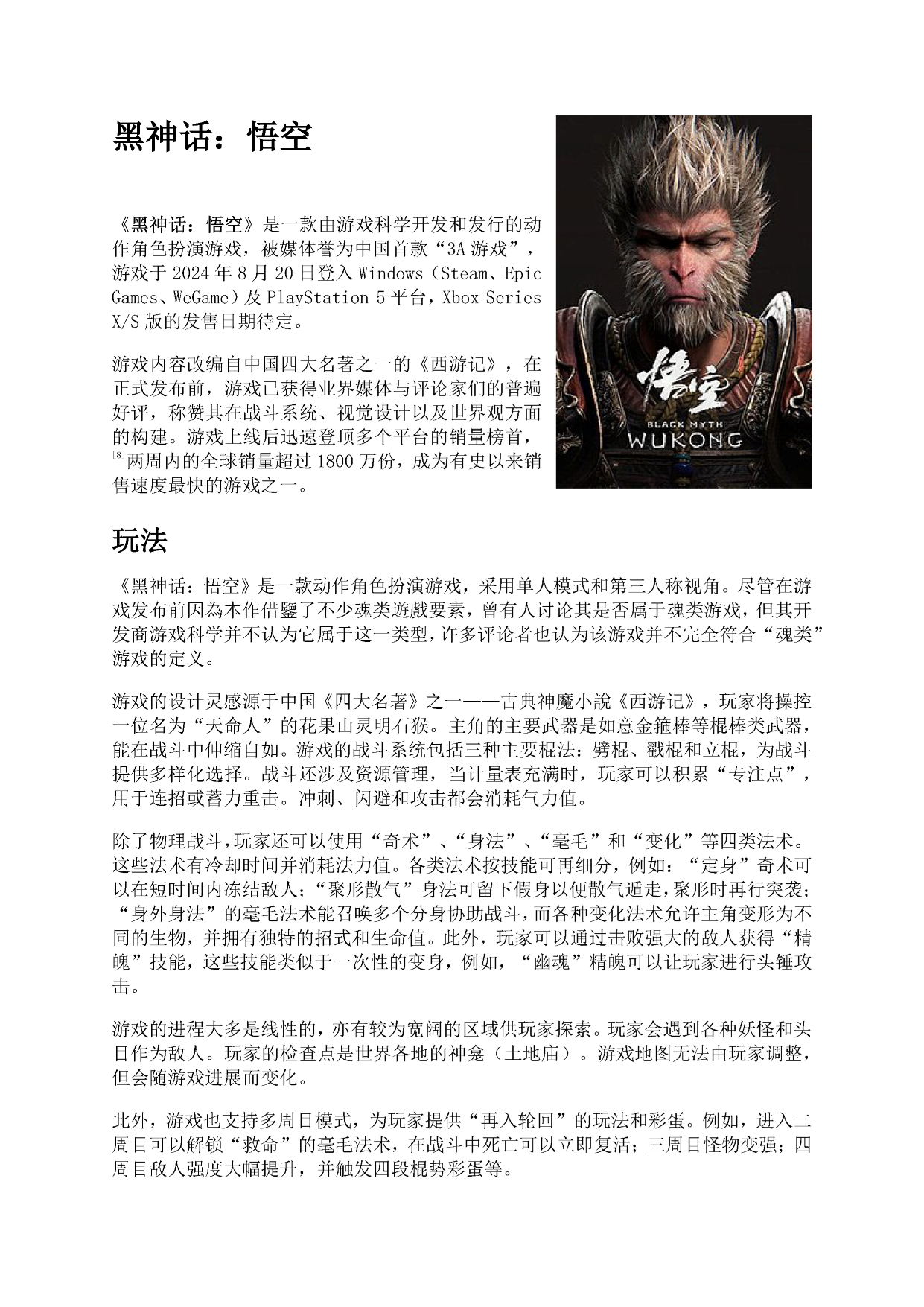}}
    \label{fig:cn_img}
  }
  \subfigure[]{
    \fbox{\includegraphics[height=1100\in]{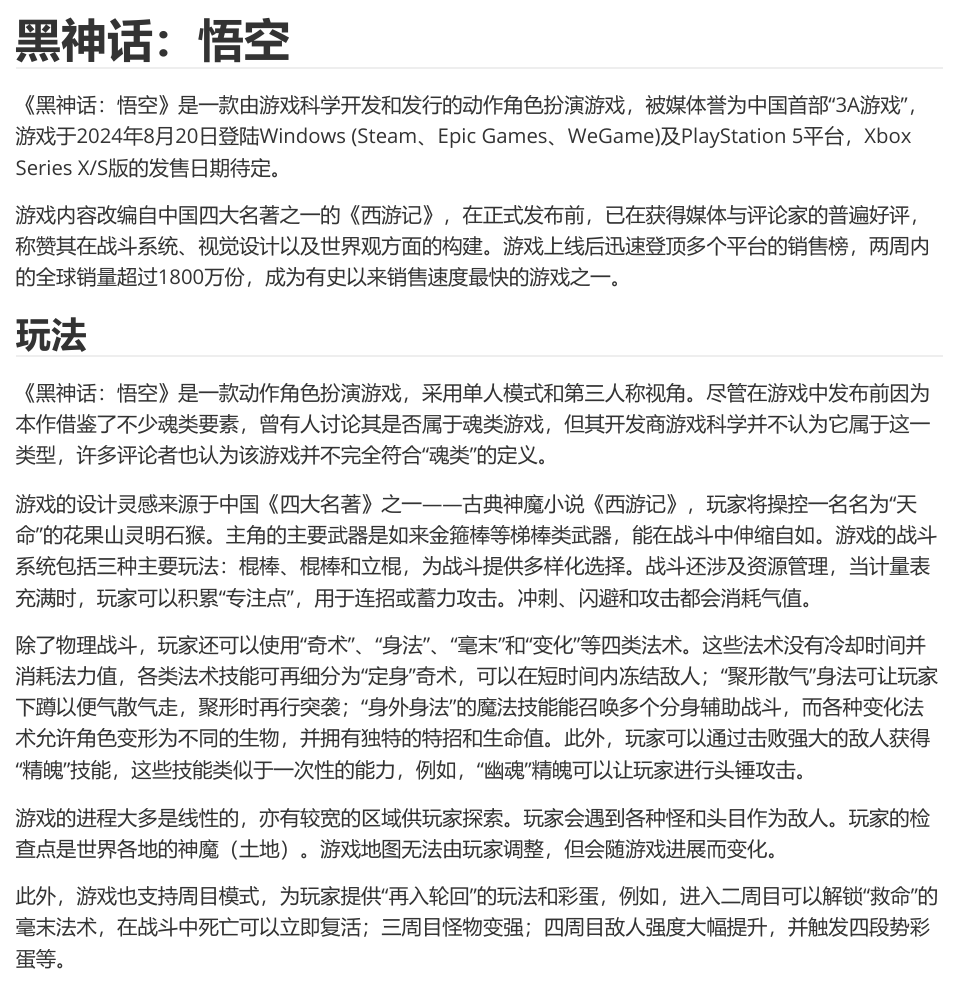}}
    \label{fig:cn_md}
  }
  \caption{Examples of image-to-markdown.}
  \label{fig:demo_markdown}
\end{figure}

\subsection{Markdown Converter}
\label{ssec:vis_ocr}
\modelname demonstrates a strong capability to transcribe content from screenshots of scientific papers, README files, and DOCX documents into markdown text. Two examples are illustrated in Fig.~\ref{fig:demo_markdown}. Notably, in the first example, \modelname accurately extracts plain text as well as precisely captures \LaTeX~formulas for complex mathematical expressions. This demonstrates the potential of LVLMs over traditional OCR engines in layout-aware OCR tasks. Additionally, the second example shows that despite the initial visual encoder not being pre-trained on a Chinese corpus, \modelname still achieves impressive Chinese OCR performance through LVLM joint training. However, a challenge for Chinese OCR still stands that LVLMs are short at recognizing uncommon words. For example, in the second last paragraph, the Chinese word ``\begin{CJK}{UTF8}{gbsn}\xpinyin*{神龛}\end{CJK}''~(shrine) is mistakenly recognized as ``\begin{CJK}{UTF8}{gbsn}\xpinyin*{神魔}\end{CJK}''~(gods and demons), which are similar in shape but significantly different in meaning. To solve these problems, greater emphasis should be placed on improving the training strategy and refining the OCR data in future work.

\begin{figure}[t]
  \centering
  \includegraphics[width=\linewidth]{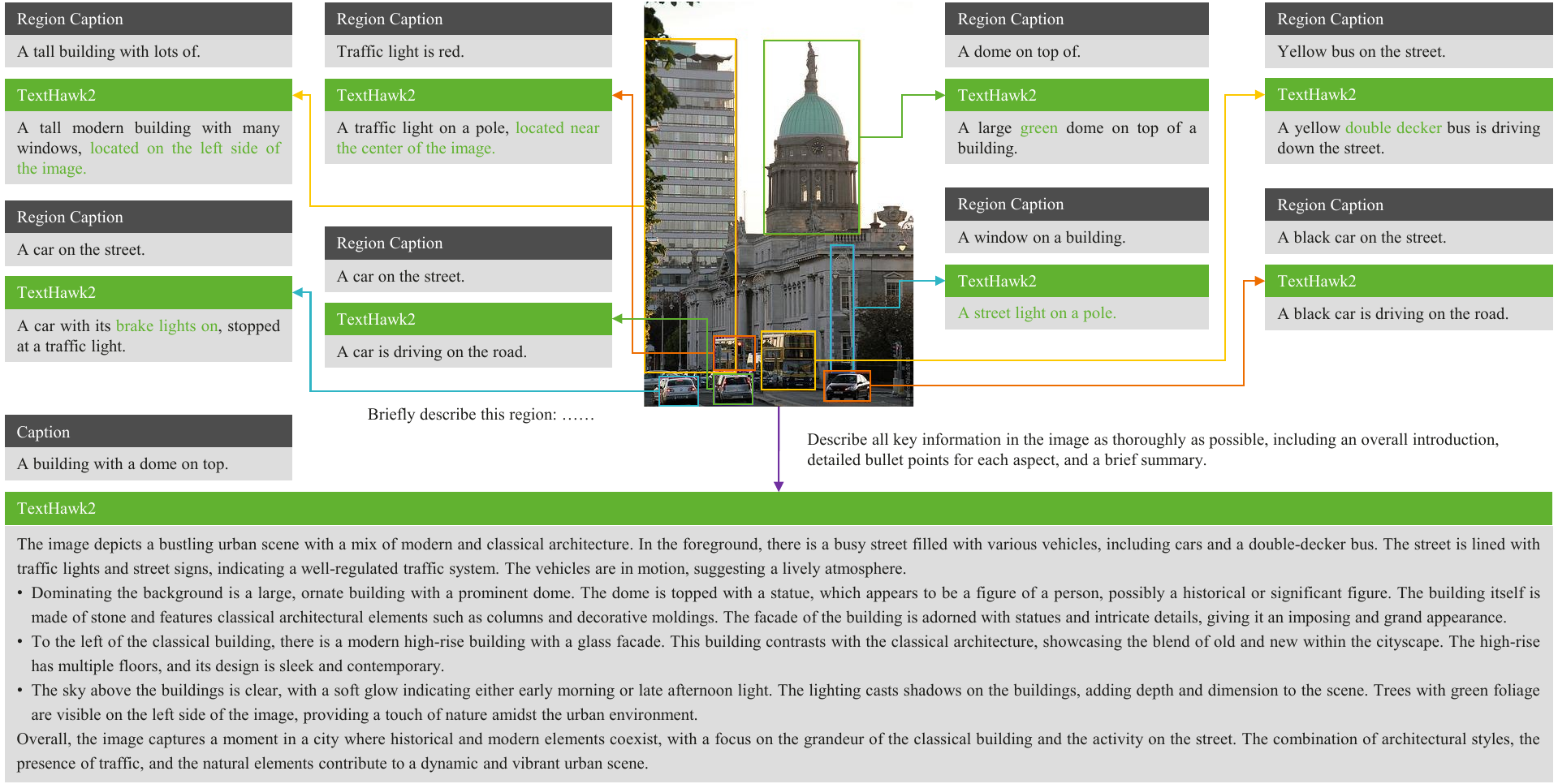}
  \caption{Comparison of captions from the UMG-41M dataset, along with those re-generated by \modelname. The bounding boxes are provided in the text prompts but are not visible in the images.}
  \label{fig:reg}
\end{figure}

\subsection{Grounding Captioner}
\label{ssec:reg}
Most large-scale visual grounding datasets, such as grounding captions and referring expressions, are generated using outdated specialist models or region-based captioners like GLIP~\citep{glip} and GPT4RoI~\citep{gpt4roi}. However, data quality plays a critical role in the final performance of LVLMs and often becomes a bottleneck for training more advanced models. Unlike generic image captions, which can be widely collected from the web and recaptioned by proprietary commercial APIs or open-source LVLMs, visual grounding data are much harder to generate and remain under-explored. This is primarily because current proprietary models lack strong grounding capabilities, and there is no high-performing open-source foundational LVLM specifically designed for grounding tasks. One potential solution for data augmentation of visual grounding involves a data-model-iteration loop, utilizing smaller detection models. A similar approach has been explored in VILA$^2$~\citep{vila2}, which introduces the concepts of self-augment and specialist-augment. In the specialist-augment step, an LVLM is fine-tuned on a high-quality subset of grounding captions and then used to recaption the remaining large-scale image dataset. It has been shown that image caption quality improves across up to three iterations.

To assess the effectiveness of \modelname as a grounding captioner, we compare the original region captions from UMG-41M with those generated by \modelname, which uses bounding boxes as additional inputs, as shown in Fig.~\ref{fig:reg}. Although \modelname is pre-trained on UMG-41M, its re-generated captions provide more detailed descriptions and better spatial relationships compared to the original captions. Unlike the specialist-augment method from VILA$^2$, our approach integrates detection results from convolutional models, which produce more accurate bounding boxes. We believe this strategy can help address distribution issues that arise in data augmentation loops, and we plan to explore this direction further in future work.

\subsection{Comparison with Proprietary Models}
\label{ssec:closed-source}
While \modelname is designed for computational efficiency and optimized for fine-grained tasks, it also demonstrates strong performance on general VQA tasks. We conduct a comprehensive comparison with proprietary models across various benchmarks, including general multimodal understanding and OCR tasks. Grounding tasks are not shown here since they have limited support. The benchmarks we consider are MMMU~\citep{mmmu}, MMBench~\citep{mmbench}, MME~\citep{mme}, MMStar~\citep{mmstar}, BLINK~\citep{blink}, MMT-Bench~\citep{mmtbench}, RealWorldQA~\citep{rwqa}, SEED-Bench~\citep{seedbench}, AI2D~\citep{ai2d}, ScienceQA~\citep{sqa}, MathVista~\citep{mathvista}, HallusionBench~\citep{hallbench}, TextVQA~\citep{textvqa}, OCRBench~\citep{ocrbench}, ChartQA~\citep{chartqa}, and DocVQA~\citep{docvqa}. As illustrated in Table~\ref{tab:benchmark}, \modelname achieves competitive results with similar-scale closed-source models in most benchmarks. Notably, \modelname scores 49.5\% on HallusionBench, despite lacking supervision from reinforcement learning methods like RLHF-V~\citep{rlhf-v}. This suggests that training on grounding tasks may help reduce hallucinations. The most significant gap between \modelname and GPT-4o-mini is observed on MMMU, suggesting that \modelname has limitations in advanced and complex tasks. This is likely due to insufficient knowledge and reasoning data and the disparity between foundational LLMs. In text-oriented benchmarks, \modelname either matches or surpasses state-of-the-art models, which utilize OCR engines or Chain-of-Thought~(CoT) prompting during inference.

\renewcommand{\thefootnote}{\fnsymbol{footnote}}
\begin{table}[t]
  \small
  \centering
  \caption{Performance comparison with proprietary models on vision-language benchmarks. Results are evaluated in VLMEvalKit~\citep{vlmevalkit} using official APIs by default.}
  \label{tab:benchmark}
  \setlength{\tabcolsep}{1.3mm}\begin{tabular}{llllll}
    \toprule
    \textbf{Benchmark} & \textbf{GPT-4o-mini} & \textbf{Gemini-1.5-Flash} & \textbf{Claude3-Haiku} & \textbf{\modelname} \\
    \toprule
    MMMU$_\text{val}$~\citep{mmmu}                  & \textbf{60.0} & \underline{58.2} & 49.7 & 45.0 \\
    MMBench-1.1$_\text{test-EN}$~\citep{mmbench}    & \textbf{77.1} & \textbf{77.1} & 58.0 & 75.0 \\
    MMBench-1.1$_\text{test-CN}$~\citep{mmbench}    & 75.0 & \textbf{76.7} & 56.2 & \underline{75.6} \\
    MMBench$_\text{test-EN}$~\citep{mmbench}        & \underline{77.6} & \textbf{79.4} & 60.7 & 77.5 \\
    MMBench$_\text{test-CN}$~\citep{mmbench}        & 75.9 & \textbf{78.6} & 57.2 & \underline{77.6} \\
    MME~\citep{mme}                                 & 2003.4 & \underline{2077.9} & 1453.2 & \textbf{2125.9} \\
    MMStar~\citep{mmstar}                           & \underline{54.8} & \textbf{55.8} & 38.1 & 54.5 \\
    BLINK~\citep{blink}                             & \underline{53.6} & \textbf{57.7} & 37.5 & 48.7 \\
    MMT-Bench$_\text{val}$~\citep{mmtbench}         & \underline{61.2} & \textbf{62.6} & 50.0 & 56.4 \\
    RealWorldQA~\citep{rwqa}                        & \underline{67.1} & \textbf{69.0} & 45.5 & 66.8 \\
    SEED-Bench$_\text{img}$~\citep{seedbench}       & 72.8 & \textbf{75.0} & 63.3 & \underline{74.3} \\
    AI2D$_\text{test}$~\citep{ai2d}                 & \underline{77.8} & \textbf{78.5} & 65.6 & 75.7 \\
    ScienceQA$_\text{test}$~\citep{sqa}             & \underline{85.4} & 83.3 & - & \textbf{85.8} \\
    MathVista$_\text{testmini}$~\citep{mathvista}   & \underline{52.4} & 51.2 & 42.2 & \textbf{54.5}\\
    HallusionBench~\citep{hallbench}                & 46.1 & \underline{48.5} & 39.2 & \textbf{49.5} \\
    \midrule
    TextVQA$_\text{val}$~\citep{textvqa}            & - & \textbf{78.7\footnotemark[1]} & - & \underline{76.1} \\
    OCRBench~\citep{ocrbench}                       & \textbf{78.5} & 75.3 & 65.8 & \underline{78.4} \\
    ChartQA$_\text{test}$~\citep{chartqa}           & 26.3 & \textcolor{gray}{\textbf{85.4}\footnotemark[1]\footnotemark[2]} & \textcolor{gray}{81.7\footnotemark[1]\footnotemark[2]} & \textbf{81.4} \\
    DocVQA$_\text{test}$~\citep{docvqa}             & 70.1 & \textcolor{gray}{\textbf{89.9}\footnotemark[1]\footnotemark[3]} & 88.8 & \textbf{89.6} \\
    \bottomrule
  \end{tabular}
\end{table}
\footnotetext[1]{indicates results from official reports.}
\footnotetext[2]{indicates Chain-of-Thought prompting.}
\footnotetext[3]{indicates using extra annotations from OCR engine.}
\renewcommand{\thefootnote}{\arabic{footnote}}

\section{Conclusion and Limitations}

In this work, we address two key questions: \textit{Can we increase the compression ratio to 16 without losing the ability to perceive fine-grained details and achieve state-of-the-art OCR performance with limited resources?} And \textit{can we train an LVLM with a single visual encoder that excels in general multimodal understanding, OCR, and grounding simultaneously?} To answer these, we introduce \modelname, which demonstrates state-of-the-art performance in multimodal understanding, OCR, and grounding, all while achieving a 16 times token compression ratio with a unified visual encoder. Notably, \modelname is pre-trained on a relatively modest dataset of 100 million samples---fewer than comparable LVLMs---highlighting the significance of visual encoder reinforcement and data diversity. Meanwhile, we optimize the data pipeline and model parallelism to boost training throughput, allowing \modelname to be trained using limited resources.

However, our experiments face several limitations. First, the training data contains insufficient scene text, limiting the model's ability to accurately recognize complex Chinese characters. Second, the supervised fine-tuning process lacks adequate multimodal knowledge and reasoning data, which affects performance in these areas. Third, the potential of native resolution ViT and full-parameter pre-training remains unexplored. Lastly, the current version of \modelname does not incorporate Reinforcement Learning from Human Feedback~(RLHF), which could help reduce hallucinations. Addressing these limitations will be essential in future work. 

\bibliography{iclr2025_conference}
\bibliographystyle{iclr2025_conference}


\end{document}

%% file: math_commands.tex
\usepackage{amsmath,amsfonts,bm}

\def\eqref#1{equation~\ref{#1}}

\def\1{\bm{1}}

\def\ve{{\bm{e}}}

\DeclareMathAlphabet{\mathsfit}{\encodingdefault}{\sfdefault}{m}{sl}
\SetMathAlphabet{\mathsfit}{bold}{\encodingdefault}{\sfdefault}{bx}{n}

\def\sR{{\mathbb{R}}}










%% file: iclr2025_conference.bbl
\begin{thebibliography}{137}
\providecommand{\natexlab}[1]{#1}
\providecommand{\url}[1]{\texttt{#1}}
\expandafter\ifx\csname urlstyle\endcsname\relax
  \providecommand{\doi}[1]{doi: #1}\else
  \providecommand{\doi}{doi: \begingroup \urlstyle{rm}\Url}\fi

\bibitem[AI@Meta(2024)]{llama3}
AI@Meta.
\newblock Llama 3 model card.
\newblock \url{https://github.com/meta-llama/llama3/blob/main/MODEL_CARD.md}, 2024.

\bibitem[Alayrac et~al.(2022)Alayrac, Donahue, Luc, Miech, Barr, Hasson, Lenc, Mensch, Millican, Reynolds, Ring, Rutherford, Cabi, Han, Gong, Samangooei, Monteiro, Menick, Borgeaud, Brock, Nematzadeh, Sharifzadeh, Binkowski, Barreira, Vinyals, Zisserman, and Simonyan]{flamingo}
Jean{-}Baptiste Alayrac, Jeff Donahue, Pauline Luc, Antoine Miech, Iain Barr, Yana Hasson, Karel Lenc, Arthur Mensch, Katherine Millican, Malcolm Reynolds, Roman Ring, Eliza Rutherford, Serkan Cabi, Tengda Han, Zhitao Gong, Sina Samangooei, Marianne Monteiro, Jacob~L. Menick, Sebastian Borgeaud, Andy Brock, Aida Nematzadeh, Sahand Sharifzadeh, Mikolaj Binkowski, Ricardo Barreira, Oriol Vinyals, Andrew Zisserman, and Kar{\'{e}}n Simonyan.
\newblock Flamingo: a visual language model for few-shot learning.
\newblock In \emph{NeurIPS}, 2022.

\bibitem[Ascend(2024)]{npu_fusion_attention}
Ascend.
\newblock Ascend extension for pytorch.
\newblock \url{https://www.hiascend.com/document/detail/zh/Pytorch/60RC2/apiref/apilist/ptaoplist_000142.html}, 2024.

\bibitem[Bai et~al.(2023)Bai, Bai, Yang, Wang, Tan, Wang, Lin, Zhou, and Zhou]{qwen-vl}
Jinze Bai, Shuai Bai, Shusheng Yang, Shijie Wang, Sinan Tan, Peng Wang, Junyang Lin, Chang Zhou, and Jingren Zhou.
\newblock Qwen-vl: {A} frontier large vision-language model with versatile abilities.
\newblock \emph{CoRR}, abs/2308.12966, 2023.

\bibitem[Bai et~al.(2024)Bai, Du, Liang, Jin, Liu, Zhou, Zheng, Zhang, Ma, Wang, Yuan, Wu, Lin, Huang, Zhang, Chen, Lin, Fu, Yang, Ni, and Zhang]{coig_cqia}
Yuelin Bai, Xinrun Du, Yiming Liang, Yonggang Jin, Ziqiang Liu, Junting Zhou, Tianyu Zheng, Xincheng Zhang, Nuo Ma, Zekun Wang, Ruibin Yuan, Haihong Wu, Hongquan Lin, Wenhao Huang, Jiajun Zhang, Wenhu Chen, Chenghua Lin, Jie Fu, Min Yang, Shiwen Ni, and Ge~Zhang.
\newblock {COIG-CQIA:} quality is all you need for chinese instruction fine-tuning.
\newblock \emph{CoRR}, abs/2403.18058, 2024.

\bibitem[Biten et~al.(2019)Biten, Tito, Mafla, i~Bigorda, Rusi{\~{n}}ol, Jawahar, Valveny, and Karatzas]{stvqa}
Ali~Furkan Biten, Rub{\`{e}}n Tito, Andr{\'{e}}s Mafla, Llu{\'{\i}}s~G{\'{o}}mez i~Bigorda, Mar{\c{c}}al Rusi{\~{n}}ol, C.~V. Jawahar, Ernest Valveny, and Dimosthenis Karatzas.
\newblock Scene text visual question answering.
\newblock In \emph{ICCV}, 2019.

\bibitem[Cai et~al.(2024)Cai, Cao, Chen, Chen, Chen, Chen, Chen, Chen, Chen, Chu, Dong, Duan, Fan, Fei, Gao, Ge, Gu, Gu, Gui, Guo, Guo, He, Hu, Huang, Jiang, Jiao, Jin, Lei, Li, Li, Li, Li, Li, Li, Liu, Liu, Hong, Liu, Liu, Liu, Lv, Lv, Lv, Ma, Ma, Ma, Ning, Ouyang, Qiu, Qu, Shang, Shao, Song, Song, Sui, Sun, Sun, Tang, Wang, Wang, Wang, Wang, Wang, Wang, Wang, Wei, Weng, Wu, Xiong, and et~al.]{internlm2}
Zheng Cai, Maosong Cao, Haojiong Chen, Kai Chen, Keyu Chen, Xin Chen, Xun Chen, Zehui Chen, Zhi Chen, Pei Chu, Xiaoyi Dong, Haodong Duan, Qi~Fan, Zhaoye Fei, Yang Gao, Jiaye Ge, Chenya Gu, Yuzhe Gu, Tao Gui, Aijia Guo, Qipeng Guo, Conghui He, Yingfan Hu, Ting Huang, Tao Jiang, Penglong Jiao, Zhenjiang Jin, Zhikai Lei, Jiaxing Li, Jingwen Li, Linyang Li, Shuaibin Li, Wei Li, Yining Li, Hongwei Liu, Jiangning Liu, Jiawei Hong, Kaiwen Liu, Kuikun Liu, Xiaoran Liu, Chengqi Lv, Haijun Lv, Kai Lv, Li~Ma, Runyuan Ma, Zerun Ma, Wenchang Ning, Linke Ouyang, Jiantao Qiu, Yuan Qu, Fukai Shang, Yunfan Shao, Demin Song, Zifan Song, Zhihao Sui, Peng Sun, Yu~Sun, Huanze Tang, Bin Wang, Guoteng Wang, Jiaqi Wang, Jiayu Wang, Rui Wang, Yudong Wang, Ziyi Wang, Xingjian Wei, Qizhen Weng, Fan Wu, Yingtong Xiong, and et~al.
\newblock Internlm2 technical report.
\newblock \emph{CoRR}, abs/2403.17297, 2024.

\bibitem[Cao et~al.(2024)Cao, Ye, Li, Yu, Tang, Lu, and Chen]{madtp}
Jianjian Cao, Peng Ye, Shengze Li, Chong Yu, Yansong Tang, Jiwen Lu, and Tao Chen.
\newblock {MADTP:} multimodal alignment-guided dynamic token pruning for accelerating vision-language transformer.
\newblock \emph{CoRR}, abs/2403.02991, 2024.

\bibitem[Changpinyo et~al.(2021)Changpinyo, Sharma, Ding, and Soricut]{cc12m}
Soravit Changpinyo, Piyush Sharma, Nan Ding, and Radu Soricut.
\newblock Conceptual 12m: Pushing web-scale image-text pre-training to recognize long-tail visual concepts.
\newblock In \emph{CVPR}, 2021.

\bibitem[Chen et~al.(2023)Chen, Zhang, Zeng, Zhang, Zhu, and Zhao]{shikra}
Keqin Chen, Zhao Zhang, Weili Zeng, Richong Zhang, Feng Zhu, and Rui Zhao.
\newblock Shikra: Unleashing multimodal llm's referential dialogue magic.
\newblock \emph{CoRR}, abs/2306.15195, 2023.

\bibitem[Chen et~al.(2024{\natexlab{a}})Chen, Li, Dong, Zhang, Zang, Chen, Duan, Wang, Qiao, Lin, and Zhao]{mmstar}
Lin Chen, Jinsong Li, Xiaoyi Dong, Pan Zhang, Yuhang Zang, Zehui Chen, Haodong Duan, Jiaqi Wang, Yu~Qiao, Dahua Lin, and Feng Zhao.
\newblock Are we on the right way for evaluating large vision-language models?
\newblock \emph{CoRR}, abs/2403.20330, 2024{\natexlab{a}}.

\bibitem[Chen et~al.(2020)Chen, Wang, Chen, Zhang, Wang, Li, Zhou, and Wang]{tabfact}
Wenhu Chen, Hongmin Wang, Jianshu Chen, Yunkai Zhang, Hong Wang, Shiyang Li, Xiyou Zhou, and William~Yang Wang.
\newblock Tabfact: {A} large-scale dataset for table-based fact verification.
\newblock In \emph{ICLR}, 2020.

\bibitem[Chen et~al.(2024{\natexlab{b}})Chen, Wang, Tian, Ye, Gao, Cui, Tong, Hu, Luo, Ma, Ma, Wang, Dong, Yan, Guo, He, Shi, Jin, Xu, Wang, Wei, Li, Zhang, Zhang, Cai, Wen, Yan, Dou, Lu, Zhu, Lu, Lin, Qiao, Dai, and Wang]{internvl-1.5}
Zhe Chen, Weiyun Wang, Hao Tian, Shenglong Ye, Zhangwei Gao, Erfei Cui, Wenwen Tong, Kongzhi Hu, Jiapeng Luo, Zheng Ma, Ji~Ma, Jiaqi Wang, Xiaoyi Dong, Hang Yan, Hewei Guo, Conghui He, Botian Shi, Zhenjiang Jin, Chao Xu, Bin Wang, Xingjian Wei, Wei Li, Wenjian Zhang, Bo~Zhang, Pinlong Cai, Licheng Wen, Xiangchao Yan, Min Dou, Lewei Lu, Xizhou Zhu, Tong Lu, Dahua Lin, Yu~Qiao, Jifeng Dai, and Wenhai Wang.
\newblock How far are we to gpt-4v? closing the gap to commercial multimodal models with open-source suites.
\newblock \emph{CoRR}, abs/2404.16821, 2024{\natexlab{b}}.

\bibitem[Chen et~al.(2024{\natexlab{c}})Chen, Wang, Tian, Ye, Gao, Cui, Tong, Hu, Luo, Ma, et~al.]{internvl2}
Zhe Chen, Weiyun Wang, Hao Tian, Shenglong Ye, Zhangwei Gao, Erfei Cui, Wenwen Tong, Kongzhi Hu, Jiapeng Luo, Zheng Ma, et~al.
\newblock Internvl2: Better than the best—expanding performance boundaries of open-source multimodal models with the progressive scaling strategy.
\newblock \url{https://internvl.github.io/blog/2024-07-02-InternVL-2.0}, 2024{\natexlab{c}}.

\bibitem[Chng et~al.(2019)Chng, Ding, Liu, Karatzas, Chan, Jin, Liu, Sun, Ng, Luo, Ni, Fang, Zhang, and Han]{art}
Chee~Kheng Chng, Errui Ding, Jingtuo Liu, Dimosthenis Karatzas, Chee~Seng Chan, Lianwen Jin, Yuliang Liu, Yipeng Sun, Chun~Chet Ng, Canjie Luo, Zihan Ni, ChuanMing Fang, Shuaitao Zhang, and Junyu Han.
\newblock {ICDAR2019} robust reading challenge on arbitrary-shaped text - rrc-art.
\newblock In \emph{ICDAR}, 2019.

\bibitem[DeepSeek{-}AI et~al.(2024)DeepSeek{-}AI, Liu, Feng, Wang, Wang, Liu, Zhao, Deng, Ruan, Dai, Guo, Yang, Chen, Ji, Li, Lin, Luo, Hao, Chen, Li, Zhang, Xu, Yang, Zhang, Ding, Xin, Gao, Li, Qu, Cai, Liang, Guo, Ni, Li, Chen, Yuan, Qiu, Song, Dong, Gao, Guan, Wang, Zhang, Xu, Xia, Zhao, Zhang, Li, Wang, Zhang, Zhang, Tang, Li, Tian, Huang, Wang, Zhang, Zhu, Chen, Du, Chen, Jin, Ge, Pan, Xu, Chen, Li, Lu, Zhou, Chen, Wu, Ye, Ma, Wang, Zhou, Yu, Zhou, Zheng, Wang, Pei, Yuan, Sun, Xiao, Zeng, An, Liu, Liang, Gao, Zhang, Li, Jin, Wang, Bi, Liu, Wang, Shen, Chen, Chen, Nie, and Sun]{deepseek-v2}
DeepSeek{-}AI, Aixin Liu, Bei Feng, Bin Wang, Bingxuan Wang, Bo~Liu, Chenggang Zhao, Chengqi Deng, Chong Ruan, Damai Dai, Daya Guo, Dejian Yang, Deli Chen, Dongjie Ji, Erhang Li, Fangyun Lin, Fuli Luo, Guangbo Hao, Guanting Chen, Guowei Li, Hao Zhang, Hanwei Xu, Hao Yang, Haowei Zhang, Honghui Ding, Huajian Xin, Huazuo Gao, Hui Li, Hui Qu, J.~L. Cai, Jian Liang, Jianzhong Guo, Jiaqi Ni, Jiashi Li, Jin Chen, Jingyang Yuan, Junjie Qiu, Junxiao Song, Kai Dong, Kaige Gao, Kang Guan, Lean Wang, Lecong Zhang, Lei Xu, Leyi Xia, Liang Zhao, Liyue Zhang, Meng Li, Miaojun Wang, Mingchuan Zhang, Minghua Zhang, Minghui Tang, Mingming Li, Ning Tian, Panpan Huang, Peiyi Wang, Peng Zhang, Qihao Zhu, Qinyu Chen, Qiushi Du, R.~J. Chen, R.~L. Jin, Ruiqi Ge, Ruizhe Pan, Runxin Xu, Ruyi Chen, S.~S. Li, Shanghao Lu, Shangyan Zhou, Shanhuang Chen, Shaoqing Wu, Shengfeng Ye, Shirong Ma, Shiyu Wang, Shuang Zhou, Shuiping Yu, Shunfeng Zhou, Size Zheng, Tao Wang, Tian Pei, Tian Yuan, Tianyu Sun, W.~L. Xiao, Wangding Zeng, Wei An, Wen
  Liu, Wenfeng Liang, Wenjun Gao, Wentao Zhang, X.~Q. Li, Xiangyue Jin, Xianzu Wang, Xiao Bi, Xiaodong Liu, Xiaohan Wang, Xiaojin Shen, Xiaokang Chen, Xiaosha Chen, Xiaotao Nie, and Xiaowen Sun.
\newblock Deepseek-v2: {A} strong, economical, and efficient mixture-of-experts language model.
\newblock \emph{CoRR}, abs/2405.04434, 2024.

\bibitem[Dong et~al.(2024)Dong, Zhang, Zang, Cao, Wang, Ouyang, Zhang, Duan, Zhang, Li, Yan, Gao, Chen, Zhang, Li, Li, Wang, Chen, He, Zhang, Dai, Qiao, Lin, and Wang]{ixc2-4khd}
Xiaoyi Dong, Pan Zhang, Yuhang Zang, Yuhang Cao, Bin Wang, Linke Ouyang, Songyang Zhang, Haodong Duan, Wenwei Zhang, Yining Li, Hang Yan, Yang Gao, Zhe Chen, Xinyue Zhang, Wei Li, Jingwen Li, Wenhai Wang, Kai Chen, Conghui He, Xingcheng Zhang, Jifeng Dai, Yu~Qiao, Dahua Lin, and Jiaqi Wang.
\newblock Internlm-xcomposer2-4khd: {A} pioneering large vision-language model handling resolutions from 336 pixels to 4k {HD}.
\newblock \emph{CoRR}, abs/2404.06512, 2024.

\bibitem[Duan et~al.(2024)Duan, Yang, Qiao, Fang, Chen, Liu, Dong, Zang, Zhang, Wang, Lin, and Chen]{vlmevalkit}
Haodong Duan, Junming Yang, Yuxuan Qiao, Xinyu Fang, Lin Chen, Yuan Liu, Xiaoyi Dong, Yuhang Zang, Pan Zhang, Jiaqi Wang, Dahua Lin, and Kai Chen.
\newblock Vlmevalkit: An open-source toolkit for evaluating large multi-modality models.
\newblock \emph{CoRR}, abs/2407.11691, 2024.

\bibitem[Fan et~al.(2024)Fan, Ji, Jiang, Li, Jin, Song, Wang, Hong, Chen, Zheng, Zhang, Huang, Zheng, Xi, Zhou, Dou, Ye, Yan, Gui, Zhang, Qiu, Huang, Wu, and Jiang]{mousi}
Xiaoran Fan, Tao Ji, Changhao Jiang, Shuo Li, Senjie Jin, Sirui Song, Junke Wang, Boyang Hong, Lu~Chen, Guodong Zheng, Ming Zhang, Caishuang Huang, Rui Zheng, Zhiheng Xi, Yuhao Zhou, Shihan Dou, Junjie Ye, Hang Yan, Tao Gui, Qi~Zhang, Xipeng Qiu, Xuanjing Huang, Zuxuan Wu, and Yu{-}Gang Jiang.
\newblock Mousi: Poly-visual-expert vision-language models.
\newblock \emph{CoRR}, abs/2401.17221, 2024.

\bibitem[Fang et~al.(2024)Fang, Zhu, Lu, Wang, Molchanov, Cho, Pavone, Han, and Yin]{vila2}
Yunhao Fang, Ligeng Zhu, Yao Lu, Yan Wang, Pavlo Molchanov, Jang~Hyun Cho, Marco Pavone, Song Han, and Hongxu Yin.
\newblock Vila\({}^{\mbox{2}}\): {VILA} augmented {VILA}.
\newblock \emph{CoRR}, abs/2407.17453, 2024.

\bibitem[Fu et~al.(2023)Fu, Chen, Shen, Qin, Zhang, Lin, Qiu, Lin, Yang, Zheng, Li, Sun, and Ji]{mme}
Chaoyou Fu, Peixian Chen, Yunhang Shen, Yulei Qin, Mengdan Zhang, Xu~Lin, Zhenyu Qiu, Wei Lin, Jinrui Yang, Xiawu Zheng, Ke~Li, Xing Sun, and Rongrong Ji.
\newblock {MME:} {A} comprehensive evaluation benchmark for multimodal large language models.
\newblock \emph{CoRR}, abs/2306.13394, 2023.

\bibitem[Fu et~al.(2024)Fu, Hu, Li, Feng, Wang, Lin, Roth, Smith, Ma, and Krishna]{blink}
Xingyu Fu, Yushi Hu, Bangzheng Li, Yu~Feng, Haoyu Wang, Xudong Lin, Dan Roth, Noah~A. Smith, Wei{-}Chiu Ma, and Ranjay Krishna.
\newblock {BLINK:} multimodal large language models can see but not perceive.
\newblock \emph{CoRR}, abs/2404.12390, 2024.

\bibitem[Gao et~al.(2023)Gao, Pi, Zhang, Ye, Zhong, Wang, Hong, Han, Xu, Li, and Kong]{g-llava}
Jiahui Gao, Renjie Pi, Jipeng Zhang, Jiacheng Ye, Wanjun Zhong, Yufei Wang, Lanqing Hong, Jianhua Han, Hang Xu, Zhenguo Li, and Lingpeng Kong.
\newblock G-llava: Solving geometric problem with multi-modal large language model.
\newblock \emph{CoRR}, abs/2312.11370, 2023.

\bibitem[Gu et~al.(2022)Gu, Meng, Lu, Hou, Minzhe, Liang, Yao, Huang, Zhang, Jiang, Xu, and Xu]{wukong}
Jiaxi Gu, Xiaojun Meng, Guansong Lu, Lu~Hou, Niu Minzhe, Xiaodan Liang, Lewei Yao, Runhui Huang, Wei Zhang, Xin Jiang, Chunjing Xu, and Hang Xu.
\newblock Wukong: {A} 100 million large-scale chinese cross-modal pre-training benchmark.
\newblock In \emph{NeurIPS}, 2022.

\bibitem[Guo et~al.(2021)Guo, Feng, Yin, Xue, Mei, and Liu]{ctsu}
Yunfei Guo, Wei Feng, Fei Yin, Tao Xue, Shuqi Mei, and Cheng{-}Lin Liu.
\newblock Learning to understand traffic signs.
\newblock In \emph{ACM MM}, 2021.

\bibitem[Harlap et~al.(2018)Harlap, Narayanan, Phanishayee, Seshadri, Devanur, Ganger, and Gibbons]{pipedream}
Aaron Harlap, Deepak Narayanan, Amar Phanishayee, Vivek Seshadri, Nikhil~R. Devanur, Gregory~R. Ganger, and Phillip~B. Gibbons.
\newblock Pipedream: Fast and efficient pipeline parallel {DNN} training.
\newblock \emph{CoRR}, abs/1806.03377, 2018.

\bibitem[He et~al.(2023)He, Jin, Xu, Qiu, Wang, Li, Yan, Wang, and Lin]{wanjuan}
Conghui He, Zhenjiang Jin, Chao Xu, Jiantao Qiu, Bin Wang, Wei Li, Hang Yan, Jiaqi Wang, and Dahua Lin.
\newblock Wanjuan: {A} comprehensive multimodal dataset for advancing english and chinese large models.
\newblock \emph{CoRR}, abs/2308.10755, 2023.

\bibitem[He et~al.(2018)He, Liu, Yang, Zhang, Luo, Gao, Zheng, Wang, Zhang, and Jin]{mtwi}
Mengchao He, Yuliang Liu, Zhibo Yang, Sheng Zhang, Canjie Luo, Feiyu Gao, Qi~Zheng, Yongpan Wang, Xin Zhang, and Lianwen Jin.
\newblock {ICPR2018} contest on robust reading for multi-type web images.
\newblock In \emph{ICPR}, 2018.

\bibitem[He et~al.(2024)He, Liu, Wu, Yuan, Wang, Huang, and Zhao]{bunny}
Muyang He, Yexin Liu, Boya Wu, Jianhao Yuan, Yueze Wang, Tiejun Huang, and Bo~Zhao.
\newblock Efficient multimodal learning from data-centric perspective.
\newblock \emph{CoRR}, abs/2402.11530, 2024.

\bibitem[Hong et~al.(2023)Hong, Wang, Lv, Xu, Yu, Ji, Wang, Wang, Zhang, Li, Xu, Dong, Ding, and Tang]{cogagent}
Wenyi Hong, Weihan Wang, Qingsong Lv, Jiazheng Xu, Wenmeng Yu, Junhui Ji, Yan Wang, Zihan Wang, Yuxuan Zhang, Juanzi Li, Bin Xu, Yuxiao Dong, Ming Ding, and Jie Tang.
\newblock Cogagent: {A} visual language model for {GUI} agents.
\newblock \emph{CoRR}, abs/2312.08914, 2023.

\bibitem[Hu et~al.(2024{\natexlab{a}})Hu, Xu, Ye, Yan, Zhang, Zhang, Li, Zhang, Jin, Huang, and Zhou]{mplug-docowl-1.5}
Anwen Hu, Haiyang Xu, Jiabo Ye, Ming Yan, Liang Zhang, Bo~Zhang, Chen Li, Ji~Zhang, Qin Jin, Fei Huang, and Jingren Zhou.
\newblock mplug-docowl 1.5: Unified structure learning for ocr-free document understanding.
\newblock \emph{CoRR}, abs/2403.12895, 2024{\natexlab{a}}.

\bibitem[Hu et~al.(2024{\natexlab{b}})Hu, Xu, Zhang, Ye, Yan, Zhang, Jin, Huang, and Zhou]{mplug-docowl-2}
Anwen Hu, Haiyang Xu, Liang Zhang, Jiabo Ye, Ming Yan, Ji~Zhang, Qin Jin, Fei Huang, and Jingren Zhou.
\newblock mplug-docowl2: High-resolution compressing for ocr-free multi-page document understanding.
\newblock \emph{CoRR}, abs/2409.03420, 2024{\natexlab{b}}.

\bibitem[Huang et~al.(2019{\natexlab{a}})Huang, Cheng, Bapna, Firat, Chen, Chen, Lee, Ngiam, Le, Wu, and Chen]{gpipe}
Yanping Huang, Youlong Cheng, Ankur Bapna, Orhan Firat, Dehao Chen, Mia~Xu Chen, HyoukJoong Lee, Jiquan Ngiam, Quoc~V. Le, Yonghui Wu, and Zhifeng Chen.
\newblock Gpipe: Efficient training of giant neural networks using pipeline parallelism.
\newblock In \emph{NeurIPS}, 2019{\natexlab{a}}.

\bibitem[Huang et~al.(2019{\natexlab{b}})Huang, Chen, He, Bai, Karatzas, Lu, and Jawahar]{sroie}
Zheng Huang, Kai Chen, Jianhua He, Xiang Bai, Dimosthenis Karatzas, Shijian Lu, and C.~V. Jawahar.
\newblock {ICDAR2019} competition on scanned receipt {OCR} and information extraction.
\newblock In \emph{ICDAR}, 2019{\natexlab{b}}.

\bibitem[Jaume et~al.(2019)Jaume, Ekenel, and Thiran]{funsd}
Guillaume Jaume, Hazim~Kemal Ekenel, and Jean{-}Philippe Thiran.
\newblock {FUNSD:} {A} dataset for form understanding in noisy scanned documents.
\newblock In \emph{OST@ICDAR}, 2019.

\bibitem[Karatzas et~al.(2015)Karatzas, Gomez{-}Bigorda, Nicolaou, Ghosh, Bagdanov, Iwamura, Matas, Neumann, Chandrasekhar, Lu, Shafait, Uchida, and Valveny]{ic15}
Dimosthenis Karatzas, Lluis Gomez{-}Bigorda, Anguelos Nicolaou, Suman~K. Ghosh, Andrew~D. Bagdanov, Masakazu Iwamura, Jiri Matas, Luk{\'{a}}s Neumann, Vijay~Ramaseshan Chandrasekhar, Shijian Lu, Faisal Shafait, Seiichi Uchida, and Ernest Valveny.
\newblock {ICDAR} 2015 competition on robust reading.
\newblock In \emph{ICDAR}, 2015.

\bibitem[Kazemzadeh et~al.(2014)Kazemzadeh, Ordonez, Matten, and Berg]{refcoco}
Sahar Kazemzadeh, Vicente Ordonez, Mark Matten, and Tamara~L. Berg.
\newblock Referitgame: Referring to objects in photographs of natural scenes.
\newblock In \emph{EMNLP}, 2014.

\bibitem[Kembhavi et~al.(2016)Kembhavi, Salvato, Kolve, Seo, Hajishirzi, and Farhadi]{ai2d}
Aniruddha Kembhavi, Mike Salvato, Eric Kolve, Min~Joon Seo, Hannaneh Hajishirzi, and Ali Farhadi.
\newblock A diagram is worth a dozen images.
\newblock In \emph{ECCV}, 2016.

\bibitem[Kirillov et~al.(2023)Kirillov, Mintun, Ravi, Mao, Rolland, Gustafson, Xiao, Whitehead, Berg, Lo, Doll{\'{a}}r, and Girshick]{sam}
Alexander Kirillov, Eric Mintun, Nikhila Ravi, Hanzi Mao, Chlo{\'{e}} Rolland, Laura Gustafson, Tete Xiao, Spencer Whitehead, Alexander~C. Berg, Wan{-}Yen Lo, Piotr Doll{\'{a}}r, and Ross~B. Girshick.
\newblock Segment anything.
\newblock In \emph{ICCV}, 2023.

\bibitem[Korthikanti et~al.(2023)Korthikanti, Casper, Lym, McAfee, Andersch, Shoeybi, and Catanzaro]{megatron3}
Vijay~Anand Korthikanti, Jared Casper, Sangkug Lym, Lawrence McAfee, Michael Andersch, Mohammad Shoeybi, and Bryan Catanzaro.
\newblock Reducing activation recomputation in large transformer models.
\newblock In \emph{MLSys}. mlsys.org, 2023.

\bibitem[Krishna et~al.(2017)Krishna, Zhu, Groth, Johnson, Hata, Kravitz, Chen, Kalantidis, Li, Shamma, Bernstein, and Fei{-}Fei]{vg}
Ranjay Krishna, Yuke Zhu, Oliver Groth, Justin Johnson, Kenji Hata, Joshua Kravitz, Stephanie Chen, Yannis Kalantidis, Li{-}Jia Li, David~A. Shamma, Michael~S. Bernstein, and Li~Fei{-}Fei.
\newblock Visual genome: Connecting language and vision using crowdsourced dense image annotations.
\newblock \emph{IJCV}, 2017.

\bibitem[Kuang et~al.(2023)Kuang, Hua, Liang, Yang, Jiang, Ren, and Bai]{poie}
Jianfeng Kuang, Wei Hua, Dingkang Liang, Mingkun Yang, Deqiang Jiang, Bo~Ren, and Xiang Bai.
\newblock Visual information extraction in the wild: Practical dataset and end-to-end solution.
\newblock In \emph{ICDAR}, 2023.

\bibitem[Lauren{\c{c}}on et~al.(2024)Lauren{\c{c}}on, Tronchon, Cord, and Sanh]{idefics2}
Hugo Lauren{\c{c}}on, L{\'{e}}o Tronchon, Matthieu Cord, and Victor Sanh.
\newblock What matters when building vision-language models?
\newblock \emph{CoRR}, abs/2405.02246, 2024.

\bibitem[Lefaudeux et~al.(2022)Lefaudeux, Massa, Liskovich, Xiong, Caggiano, Naren, Xu, Hu, Tintore, Zhang, Labatut, Haziza, Wehrstedt, Reizenstein, and Sizov]{xformers}
Benjamin Lefaudeux, Francisco Massa, Diana Liskovich, Wenhan Xiong, Vittorio Caggiano, Sean Naren, Min Xu, Jieru Hu, Marta Tintore, Susan Zhang, Patrick Labatut, Daniel Haziza, Luca Wehrstedt, Jeremy Reizenstein, and Grigory Sizov.
\newblock xformers: A modular and hackable transformer modelling library.
\newblock \url{https://github.com/facebookresearch/xformers}, 2022.

\bibitem[Lerner et~al.(2022)Lerner, Ferret, Guinaudeau, Borgne, Besan{\c{c}}on, Moreno, and Lov{\'{o}}n{-}Melgarejo]{viquae}
Paul Lerner, Olivier Ferret, Camille Guinaudeau, Herv{\'{e}}~Le Borgne, Romaric Besan{\c{c}}on, Jos{\'{e}}~G. Moreno, and Jes{\'{u}}s Lov{\'{o}}n{-}Melgarejo.
\newblock Viquae, a dataset for knowledge-based visual question answering about named entities.
\newblock In \emph{SIGIR}, 2022.

\bibitem[Lewis et~al.(2006)Lewis, Agam, Argamon, Frieder, Grossman, and Heard]{iit_cdip}
David~D. Lewis, Gady Agam, Shlomo Argamon, Ophir Frieder, David~A. Grossman, and Jefferson Heard.
\newblock Building a test collection for complex document information processing.
\newblock In \emph{ACM MM}, 2006.

\bibitem[Li et~al.(2024{\natexlab{a}})Li, Zhang, Zhang, Guo, Zhang, Li, Zhang, Liu, and Li]{llavanext-strong}
Bo~Li, Kaichen Zhang, Hao Zhang, Dong Guo, Renrui Zhang, Feng Li, Yuanhan Zhang, Ziwei Liu, and Chunyuan Li.
\newblock Llava-next: Stronger llms supercharge multimodal capabilities in the wild.
\newblock \url{https://llava-vl.github.io/blog/2024-05-10-llava-next-stronger-llms/}, 2024{\natexlab{a}}.

\bibitem[Li et~al.(2023{\natexlab{a}})Li, Wang, Wang, Ge, Ge, and Shan]{seedbench}
Bohao Li, Rui Wang, Guangzhi Wang, Yuying Ge, Yixiao Ge, and Ying Shan.
\newblock Seed-bench: Benchmarking multimodal llms with generative comprehension.
\newblock \emph{CoRR}, abs/2307.16125, 2023{\natexlab{a}}.

\bibitem[Li et~al.(2023{\natexlab{b}})Li, Li, Savarese, and Hoi]{blip2}
Junnan Li, Dongxu Li, Silvio Savarese, and Steven C.~H. Hoi.
\newblock {BLIP-2:} bootstrapping language-image pre-training with frozen image encoders and large language models.
\newblock In \emph{Proc. ICML}, 2023{\natexlab{b}}.

\bibitem[Li et~al.(2022)Li, Zhang, Zhang, Yang, Li, Zhong, Wang, Yuan, Zhang, Hwang, Chang, and Gao]{glip}
Liunian~Harold Li, Pengchuan Zhang, Haotian Zhang, Jianwei Yang, Chunyuan Li, Yiwu Zhong, Lijuan Wang, Lu~Yuan, Lei Zhang, Jenq{-}Neng Hwang, Kai{-}Wei Chang, and Jianfeng Gao.
\newblock Grounded language-image pre-training.
\newblock In \emph{CVPR}, 2022.

\bibitem[Li et~al.(2020)Li, Zhao, Varma, Salpekar, Noordhuis, Li, Paszke, Smith, Vaughan, Damania, and Chintala]{ddp}
Shen Li, Yanli Zhao, Rohan Varma, Omkar Salpekar, Pieter Noordhuis, Teng Li, Adam Paszke, Jeff Smith, Brian Vaughan, Pritam Damania, and Soumith Chintala.
\newblock Pytorch distributed: Experiences on accelerating data parallel training.
\newblock \emph{Proc. {VLDB} Endow.}, 13, 2020.

\bibitem[Li et~al.(2024{\natexlab{b}})Li, Zhang, Wang, Zhong, Chen, Chu, Liu, and Jia]{minigemini}
Yanwei Li, Yuechen Zhang, Chengyao Wang, Zhisheng Zhong, Yixin Chen, Ruihang Chu, Shaoteng Liu, and Jiaya Jia.
\newblock Mini-gemini: Mining the potential of multi-modality vision language models.
\newblock \emph{CoRR}, abs/2403.18814, 2024{\natexlab{b}}.

\bibitem[Li et~al.(2024{\natexlab{c}})Li, Luo, Zhang, Qiu, and Wei]{volcano}
Zejun Li, Ruipu Luo, Jiwen Zhang, Minghui Qiu, and Zhongyu Wei.
\newblock Vocot: Unleashing visually grounded multi-step reasoning in large multi-modal models.
\newblock \emph{CoRR}, abs/2405.16919, 2024{\natexlab{c}}.

\bibitem[Li et~al.(2023{\natexlab{c}})Li, Yang, Liu, Ma, Zhang, Yang, Sun, Liu, and Bai]{monkey}
Zhang Li, Biao Yang, Qiang Liu, Zhiyin Ma, Shuo Zhang, Jingxu Yang, Yabo Sun, Yuliang Liu, and Xiang Bai.
\newblock Monkey: Image resolution and text label are important things for large multi-modal models.
\newblock In \emph{CVPR}, 2023{\natexlab{c}}.

\bibitem[Lin et~al.(2023)Lin, Liu, Zhang, Gao, Qiu, Xiao, Qiu, Lin, Shao, Chen, Han, Huang, Zhang, He, Li, and Qiao]{sphinx}
Ziyi Lin, Chris Liu, Renrui Zhang, Peng Gao, Longtian Qiu, Han Xiao, Han Qiu, Chen Lin, Wenqi Shao, Keqin Chen, Jiaming Han, Siyuan Huang, Yichi Zhang, Xuming He, Hongsheng Li, and Yu~Qiao.
\newblock {SPHINX:} the joint mixing of weights, tasks, and visual embeddings for multi-modal large language models.
\newblock \emph{CoRR}, abs/2311.07575, 2023.

\bibitem[Liu et~al.(2024{\natexlab{a}})Liu, Yin, Cao, Jiang, Li, Liu, Jiang, Sun, and Xu]{hrvda}
Chaohu Liu, Kun Yin, Haoyu Cao, Xinghua Jiang, Xin Li, Yinsong Liu, Deqiang Jiang, Xing Sun, and Linli Xu.
\newblock {HRVDA:} high-resolution visual document assistant.
\newblock \emph{CoRR}, abs/2404.06918, 2024{\natexlab{a}}.

\bibitem[Liu et~al.(2023{\natexlab{a}})Liu, Guan, Li, Chen, Yacoob, Manocha, and Zhou]{hallbench}
Fuxiao Liu, Tianrui Guan, Zongxia Li, Lichang Chen, Yaser Yacoob, Dinesh Manocha, and Tianyi Zhou.
\newblock Hallusionbench: You see what you think? or you think what you see? an image-context reasoning benchmark challenging for gpt-4v(ision), llava-1.5, and other multi-modality models.
\newblock \emph{CoRR}, abs/2310.14566, 2023{\natexlab{a}}.

\bibitem[Liu et~al.(2024{\natexlab{b}})Liu, Wang, Yao, Chen, Song, Cho, Yacoob, and Yu]{mmc}
Fuxiao Liu, Xiaoyang Wang, Wenlin Yao, Jianshu Chen, Kaiqiang Song, Sangwoo Cho, Yaser Yacoob, and Dong Yu.
\newblock {MMC:} advancing multimodal chart understanding with large-scale instruction tuning.
\newblock In \emph{NAACL-HLT}, 2024{\natexlab{b}}.

\bibitem[Liu et~al.(2023{\natexlab{b}})Liu, Li, Wu, and Lee]{llava}
Haotian Liu, Chunyuan Li, Qingyang Wu, and Yong~Jae Lee.
\newblock Visual instruction tuning.
\newblock In \emph{NeurIPS}, 2023{\natexlab{b}}.

\bibitem[Liu et~al.(2023{\natexlab{c}})Liu, Zeng, Ren, Li, Zhang, Yang, Li, Yang, Su, Zhu, and Zhang]{ground-dino}
Shilong Liu, Zhaoyang Zeng, Tianhe Ren, Feng Li, Hao Zhang, Jie Yang, Chunyuan Li, Jianwei Yang, Hang Su, Jun Zhu, and Lei Zhang.
\newblock Grounding {DINO:} marrying {DINO} with grounded pre-training for open-set object detection.
\newblock \emph{CoRR}, abs/2303.05499, 2023{\natexlab{c}}.

\bibitem[Liu et~al.(2023{\natexlab{d}})Liu, Duan, Zhang, Li, Zhang, Zhao, Yuan, Wang, He, Liu, Chen, and Lin]{mmbench}
Yuan Liu, Haodong Duan, Yuanhan Zhang, Bo~Li, Songyang Zhang, Wangbo Zhao, Yike Yuan, Jiaqi Wang, Conghui He, Ziwei Liu, Kai Chen, and Dahua Lin.
\newblock Mmbench: Is your multi-modal model an all-around player?
\newblock \emph{CoRR}, abs/2307.06281, 2023{\natexlab{d}}.

\bibitem[Liu et~al.(2023{\natexlab{e}})Liu, Li, Li, Yu, Huang, Peng, Liu, Chen, Li, Jin, and Bai]{ocrbench}
Yuliang Liu, Zhang Li, Hongliang Li, Wenwen Yu, Mingxin Huang, Dezhi Peng, Mingyu Liu, Mingrui Chen, Chunyuan Li, Lianwen Jin, and Xiang Bai.
\newblock On the hidden mystery of {OCR} in large multimodal models.
\newblock \emph{CoRR}, abs/2305.07895, 2023{\natexlab{e}}.

\bibitem[Liu et~al.(2024{\natexlab{c}})Liu, Yang, Liu, Li, Ma, Zhang, and Bai]{textmonkey}
Yuliang Liu, Biao Yang, Qiang Liu, Zhang Li, Zhiyin Ma, Shuo Zhang, and Xiang Bai.
\newblock Textmonkey: An ocr-free large multimodal model for understanding document.
\newblock \emph{CoRR}, abs/2403.04473, 2024{\natexlab{c}}.

\bibitem[Lu et~al.(2024{\natexlab{a}})Lu, Liu, Zhang, Wang, Dong, Liu, Sun, Ren, Li, Yang, Sun, Deng, Xu, Xie, and Ruan]{deepseek-vl}
Haoyu Lu, Wen Liu, Bo~Zhang, Bingxuan Wang, Kai Dong, Bo~Liu, Jingxiang Sun, Tongzheng Ren, Zhuoshu Li, Hao Yang, Yaofeng Sun, Chengqi Deng, Hanwei Xu, Zhenda Xie, and Chong Ruan.
\newblock Deepseek-vl: Towards real-world vision-language understanding.
\newblock \emph{CoRR}, abs/2403.05525, 2024{\natexlab{a}}.

\bibitem[Lu et~al.(2022)Lu, Mishra, Xia, Qiu, Chang, Zhu, Tafjord, Clark, and Kalyan]{sqa}
Pan Lu, Swaroop Mishra, Tanglin Xia, Liang Qiu, Kai{-}Wei Chang, Song{-}Chun Zhu, Oyvind Tafjord, Peter Clark, and Ashwin Kalyan.
\newblock Learn to explain: Multimodal reasoning via thought chains for science question answering.
\newblock In \emph{NeurIPS}, 2022.

\bibitem[Lu et~al.(2024{\natexlab{b}})Lu, Bansal, Xia, Liu, Li, Hajishirzi, Cheng, Chang, Galley, and Gao]{mathvista}
Pan Lu, Hritik Bansal, Tony Xia, Jiacheng Liu, Chunyuan Li, Hannaneh Hajishirzi, Hao Cheng, Kai{-}Wei Chang, Michel Galley, and Jianfeng Gao.
\newblock Mathvista: Evaluating mathematical reasoning of foundation models in visual contexts.
\newblock In \emph{ICLR}, 2024{\natexlab{b}}.

\bibitem[Lv et~al.(2023)Lv, Huang, Chen, Cui, Ma, Chang, Huang, Wang, Dong, Luo, Wu, Wang, Zhang, and Wei]{kosmos2_5}
Tengchao Lv, Yupan Huang, Jingye Chen, Lei Cui, Shuming Ma, Yaoyao Chang, Shaohan Huang, Wenhui Wang, Li~Dong, Weiyao Luo, Shaoxiang Wu, Guoxin Wang, Cha Zhang, and Furu Wei.
\newblock Kosmos-2.5: {A} multimodal literate model.
\newblock \emph{CoRR}, abs/2309.11419, 2023.

\bibitem[Mao et~al.(2016)Mao, Huang, Toshev, Camburu, Yuille, and Murphy]{refcocog}
Junhua Mao, Jonathan Huang, Alexander Toshev, Oana Camburu, Alan~L. Yuille, and Kevin Murphy.
\newblock Generation and comprehension of unambiguous object descriptions.
\newblock In \emph{CVPR}, 2016.

\bibitem[Masry et~al.(2022)Masry, Long, Tan, Joty, and Hoque]{chartqa}
Ahmed Masry, Do~Xuan Long, Jia~Qing Tan, Shafiq~R. Joty, and Enamul Hoque.
\newblock Chartqa: {A} benchmark for question answering about charts with visual and logical reasoning.
\newblock In \emph{ACL}, 2022.

\bibitem[Mathew et~al.(2021)Mathew, Karatzas, and Jawahar]{docvqa}
Minesh Mathew, Dimosthenis Karatzas, and C.~V. Jawahar.
\newblock Docvqa: {A} dataset for {VQA} on document images.
\newblock In \emph{WACV}, pp.\  2199--2208, 2021.

\bibitem[Mathew et~al.(2022)Mathew, Bagal, Tito, Karatzas, Valveny, and Jawahar]{infovqa}
Minesh Mathew, Viraj Bagal, Rub{\`{e}}n Tito, Dimosthenis Karatzas, Ernest Valveny, and C.~V. Jawahar.
\newblock Infographicvqa.
\newblock In \emph{WACV}, 2022.

\bibitem[Meng et~al.(2024)Meng, Shao, Lu, Gao, Zhang, Qiao, and Luo]{chartast}
Fanqing Meng, Wenqi Shao, Quanfeng Lu, Peng Gao, Kaipeng Zhang, Yu~Qiao, and Ping Luo.
\newblock Chartassisstant: {A} universal chart multimodal language model via chart-to-table pre-training and multitask instruction tuning.
\newblock \emph{CoRR}, abs/2401.02384, 2024.

\bibitem[Narayanan et~al.(2021)Narayanan, Shoeybi, Casper, LeGresley, Patwary, Korthikanti, Vainbrand, Kashinkunti, Bernauer, Catanzaro, Phanishayee, and Zaharia]{megatron2}
Deepak Narayanan, Mohammad Shoeybi, Jared Casper, Patrick LeGresley, Mostofa Patwary, Vijay Korthikanti, Dmitri Vainbrand, Prethvi Kashinkunti, Julie Bernauer, Bryan Catanzaro, Amar Phanishayee, and Matei Zaharia.
\newblock Efficient large-scale language model training on {GPU} clusters using megatron-lm.
\newblock In \emph{SC}, 2021.

\bibitem[Nayef et~al.(2019)Nayef, Liu, Ogier, Patel, Busta, Chowdhury, Karatzas, Khlif, Matas, Pal, and Burie]{mlt}
Nibal Nayef, Cheng{-}Lin Liu, Jean{-}Marc Ogier, Yash Patel, Michal Busta, Pinaki~Nath Chowdhury, Dimosthenis Karatzas, Wafa Khlif, Jiri Matas, Umapada Pal, and Jean{-}Christophe Burie.
\newblock {ICDAR2019} robust reading challenge on multi-lingual scene text detection and recognition - {RRC-MLT-2019}.
\newblock In \emph{ICDAR}, 2019.

\bibitem[Oquab et~al.(2023)Oquab, Darcet, Moutakanni, Vo, Szafraniec, Khalidov, Fernandez, Haziza, Massa, El{-}Nouby, Assran, Ballas, Galuba, Howes, Huang, Li, Misra, Rabbat, Sharma, Synnaeve, Xu, J{\'{e}}gou, Mairal, Labatut, Joulin, and Bojanowski]{dinov2}
Maxime Oquab, Timoth{\'{e}}e Darcet, Th{\'{e}}o Moutakanni, Huy Vo, Marc Szafraniec, Vasil Khalidov, Pierre Fernandez, Daniel Haziza, Francisco Massa, Alaaeldin El{-}Nouby, Mahmoud Assran, Nicolas Ballas, Wojciech Galuba, Russell Howes, Po{-}Yao Huang, Shang{-}Wen Li, Ishan Misra, Michael~G. Rabbat, Vasu Sharma, Gabriel Synnaeve, Hu~Xu, Herv{\'{e}} J{\'{e}}gou, Julien Mairal, Patrick Labatut, Armand Joulin, and Piotr Bojanowski.
\newblock Dinov2: Learning robust visual features without supervision.
\newblock \emph{CoRR}, abs/2304.07193, 2023.

\bibitem[Ordonez et~al.(2011)Ordonez, Kulkarni, and Berg]{sbu}
Vicente Ordonez, Girish Kulkarni, and Tamara~L. Berg.
\newblock Im2text: Describing images using 1 million captioned photographs.
\newblock In \emph{NeurIPS}, 2011.

\bibitem[Pasupat \& Liang(2015)Pasupat and Liang]{wtq}
Panupong Pasupat and Percy Liang.
\newblock Compositional semantic parsing on semi-structured tables.
\newblock In \emph{ACL}, 2015.

\bibitem[Peng et~al.(2023)Peng, Wang, Dong, Hao, Huang, Ma, and Wei]{kosmos2}
Zhiliang Peng, Wenhui Wang, Li~Dong, Yaru Hao, Shaohan Huang, Shuming Ma, and Furu Wei.
\newblock Kosmos-2: Grounding multimodal large language models to the world.
\newblock \emph{CoRR}, abs/2306.14824, 2023.

\bibitem[Radford et~al.(2021)Radford, Kim, Hallacy, Ramesh, Goh, Agarwal, Sastry, Askell, Mishkin, Clark, Krueger, and Sutskever]{clip}
Alec Radford, Jong~Wook Kim, Chris Hallacy, Aditya Ramesh, Gabriel Goh, Sandhini Agarwal, Girish Sastry, Amanda Askell, Pamela Mishkin, Jack Clark, Gretchen Krueger, and Ilya Sutskever.
\newblock Learning transferable visual models from natural language supervision.
\newblock In \emph{Proc. ICML}, volume 139, 2021.

\bibitem[Rajbhandari et~al.(2020)Rajbhandari, Rasley, Ruwase, and He]{zero}
Samyam Rajbhandari, Jeff Rasley, Olatunji Ruwase, and Yuxiong He.
\newblock Zero: memory optimizations toward training trillion parameter models.
\newblock In \emph{SC}, 2020.

\bibitem[Ridnik et~al.(2021)Ridnik, Baruch, Noy, and Zelnik]{in21k}
Tal Ridnik, Emanuel~Ben Baruch, Asaf Noy, and Lihi Zelnik.
\newblock Imagenet-21k pretraining for the masses.
\newblock In \emph{NeurIPS}, 2021.

\bibitem[Schuhmann et~al.(2021)Schuhmann, Vencu, Beaumont, Kaczmarczyk, Mullis, Katta, Coombes, Jitsev, and Komatsuzaki]{laion400m}
Christoph Schuhmann, Richard Vencu, Romain Beaumont, Robert Kaczmarczyk, Clayton Mullis, Aarush Katta, Theo Coombes, Jenia Jitsev, and Aran Komatsuzaki.
\newblock {LAION-400M:} open dataset of clip-filtered 400 million image-text pairs.
\newblock \emph{CoRR}, abs/2111.02114, 2021.

\bibitem[Shah et~al.(2019)Shah, Mishra, Yadati, and Talukdar]{kvqa}
Sanket Shah, Anand Mishra, Naganand Yadati, and Partha~Pratim Talukdar.
\newblock {KVQA:} knowledge-aware visual question answering.
\newblock In \emph{AAAI}, 2019.

\bibitem[Shang et~al.(2024)Shang, Cai, Xu, Lee, and Yan]{llavaprumerge}
Yuzhang Shang, Mu~Cai, Bingxin Xu, Yong~Jae Lee, and Yan Yan.
\newblock Llava-prumerge: Adaptive token reduction for efficient large multimodal models.
\newblock \emph{CoRR}, abs/2403.15388, 2024.

\bibitem[Sharma et~al.(2018)Sharma, Ding, Goodman, and Soricut]{cc3m}
Piyush Sharma, Nan Ding, Sebastian Goodman, and Radu Soricut.
\newblock Conceptual captions: {A} cleaned, hypernymed, image alt-text dataset for automatic image captioning.
\newblock In \emph{ACL}, 2018.

\bibitem[Shi et~al.(2017)Shi, Yao, Liao, Yang, Xu, Cui, Belongie, Lu, and Bai]{rctw17}
Baoguang Shi, Cong Yao, Minghui Liao, Mingkun Yang, Pei Xu, Linyan Cui, Serge~J. Belongie, Shijian Lu, and Xiang Bai.
\newblock {ICDAR2017} competition on reading chinese text in the wild {(RCTW-17)}.
\newblock In \emph{ICDAR}, 2017.

\bibitem[Shi et~al.(2024)Shi, Zhao, Wang, Zhang, Wang, Li, Dai, Zou, Xiong, Tian, and Zhang]{umg}
Bowen Shi, Peisen Zhao, Zichen Wang, Yuhang Zhang, Yaoming Wang, Jin Li, Wenrui Dai, Junni Zou, Hongkai Xiong, Qi~Tian, and Xiaopeng Zhang.
\newblock {UMG-CLIP:} {A} unified multi-granularity vision generalist for open-world understanding.
\newblock \emph{CoRR}, abs/2401.06397, 2024.

\bibitem[Shoeybi et~al.(2019)Shoeybi, Patwary, Puri, LeGresley, Casper, and Catanzaro]{megatron1}
Mohammad Shoeybi, Mostofa Patwary, Raul Puri, Patrick LeGresley, Jared Casper, and Bryan Catanzaro.
\newblock Megatron-lm: Training multi-billion parameter language models using model parallelism.
\newblock \emph{CoRR}, abs/1909.08053, 2019.

\bibitem[Singh et~al.(2019)Singh, Natarajan, Shah, Jiang, Chen, Batra, Parikh, and Rohrbach]{textvqa}
Amanpreet Singh, Vivek Natarajan, Meet Shah, Yu~Jiang, Xinlei Chen, Dhruv Batra, Devi Parikh, and Marcus Rohrbach.
\newblock Towards {VQA} models that can read.
\newblock In \emph{CVPR}, 2019.

\bibitem[Smock et~al.(2022)Smock, Pesala, and Abraham]{pubtables}
Brandon Smock, Rohith Pesala, and Robin Abraham.
\newblock Pubtables-1m: Towards comprehensive table extraction from unstructured documents.
\newblock In \emph{CVPR}, 2022.

\bibitem[StabilityAI \& LAION(2023)StabilityAI and LAION]{renderedtext}
StabilityAI and LAION.
\newblock Renderedtext.
\newblock \url{https://huggingface.co/datasets/wendlerc/RenderedText}, 2023.

\bibitem[Sun et~al.(2019)Sun, Karatzas, Chan, Jin, Ni, Chng, Liu, Luo, Ng, Han, Ding, and Liu]{lsvt}
Yipeng Sun, Dimosthenis Karatzas, Chee~Seng Chan, Lianwen Jin, Zihan Ni, Chee~Kheng Chng, Yuliang Liu, Canjie Luo, Chun~Chet Ng, Junyu Han, Errui Ding, and Jingtuo Liu.
\newblock {ICDAR} 2019 competition on large-scale street view text with partial labeling - {RRC-LSVT}.
\newblock In \emph{ICDAR}, 2019.

\bibitem[Teknium(2023)]{openhermes2_5}
Teknium.
\newblock Openhermes 2.5: An open dataset of synthetic data for generalist llm assistants.
\newblock \url{https://huggingface.co/datasets/teknium/OpenHermes-2.5}, 2023.

\bibitem[Thomee et~al.(2016)Thomee, Shamma, Friedland, Elizalde, Ni, Poland, Borth, and Li]{yfcc}
Bart Thomee, David~A. Shamma, Gerald Friedland, Benjamin Elizalde, Karl Ni, Douglas Poland, Damian Borth, and Li{-}Jia Li.
\newblock {YFCC100M:} the new data in multimedia research.
\newblock \emph{Commun. {ACM}}, 2016.

\bibitem[Tong et~al.(2024)Tong, Brown, Wu, Woo, Middepogu, Akula, Yang, Yang, Iyer, Pan, Wang, Fergus, LeCun, and Xie]{cambrian-1}
Shengbang Tong, Ellis Brown, Penghao Wu, Sanghyun Woo, Manoj Middepogu, Sai~Charitha Akula, Jihan Yang, Shusheng Yang, Adithya Iyer, Xichen Pan, Austin Wang, Rob Fergus, Yann LeCun, and Saining Xie.
\newblock Cambrian-1: {A} fully open, vision-centric exploration of multimodal llms.
\newblock \emph{CoRR}, abs/2406.16860, 2024.

\bibitem[Touvron et~al.(2023)Touvron, Lavril, Izacard, Martinet, Lachaux, Lacroix, Rozi{\`{e}}re, Goyal, Hambro, Azhar, Rodriguez, Joulin, Grave, and Lample]{llama}
Hugo Touvron, Thibaut Lavril, Gautier Izacard, Xavier Martinet, Marie{-}Anne Lachaux, Timoth{\'{e}}e Lacroix, Baptiste Rozi{\`{e}}re, Naman Goyal, Eric Hambro, Faisal Azhar, Aur{\'{e}}lien Rodriguez, Armand Joulin, Edouard Grave, and Guillaume Lample.
\newblock Llama: Open and efficient foundation language models.
\newblock \emph{CoRR}, abs/2302.13971, 2023.

\bibitem[Veit et~al.(2016)Veit, Matera, Neumann, Matas, and Belongie]{cocotext}
Andreas Veit, Tomas Matera, Luk{\'{a}}s Neumann, Jiri Matas, and Serge~J. Belongie.
\newblock Coco-text: Dataset and benchmark for text detection and recognition in natural images.
\newblock \emph{CoRR}, abs/1601.07140, 2016.

\bibitem[Wang et~al.(2024{\natexlab{a}})Wang, Gu, Xu, Zhang, Shi, and He]{unimer}
Bin Wang, Zhuangcheng Gu, Chao Xu, Bo~Zhang, Botian Shi, and Conghui He.
\newblock Unimernet: {A} universal network for real-world mathematical expression recognition.
\newblock \emph{CoRR}, abs/2404.15254, 2024{\natexlab{a}}.

\bibitem[Wang et~al.(2023{\natexlab{a}})Wang, Meng, Weng, He, Wu, and Jiang]{lvis_instruct4v}
Junke Wang, Lingchen Meng, Zejia Weng, Bo~He, Zuxuan Wu, and Yu{-}Gang Jiang.
\newblock To see is to believe: Prompting {GPT-4V} for better visual instruction tuning.
\newblock \emph{CoRR}, abs/2311.07574, 2023{\natexlab{a}}.

\bibitem[Wang et~al.(2022)Wang, Yang, Men, Lin, Bai, Li, Ma, Zhou, Zhou, and Yang]{ofa}
Peng Wang, An~Yang, Rui Men, Junyang Lin, Shuai Bai, Zhikang Li, Jianxin Ma, Chang Zhou, Jingren Zhou, and Hongxia Yang.
\newblock {OFA:} unifying architectures, tasks, and modalities through a simple sequence-to-sequence learning framework.
\newblock In \emph{Proc. ICML}, 2022.

\bibitem[Wang et~al.(2023{\natexlab{b}})Wang, Wang, Lin, Bai, Zhou, Zhou, Wang, and Zhou]{one-peace}
Peng Wang, Shijie Wang, Junyang Lin, Shuai Bai, Xiaohuan Zhou, Jingren Zhou, Xinggang Wang, and Chang Zhou.
\newblock {ONE-PEACE:} exploring one general representation model toward unlimited modalities.
\newblock \emph{CoRR}, abs/2305.11172, 2023{\natexlab{b}}.

\bibitem[Wang et~al.(2024{\natexlab{b}})Wang, Bai, Tan, Wang, Fan, Bai, Chen, Liu, Wang, Ge, Fan, Dang, Du, Ren, Men, Liu, Zhou, Zhou, and Lin]{qwen2-vl}
Peng Wang, Shuai Bai, Sinan Tan, Shijie Wang, Zhihao Fan, Jinze Bai, Keqin Chen, Xuejing Liu, Jialin Wang, Wenbin Ge, Yang Fan, Kai Dang, Mengfei Du, Xuancheng Ren, Rui Men, Dayiheng Liu, Chang Zhou, Jingren Zhou, and Junyang Lin.
\newblock Qwen2-vl: Enhancing vision-language model's perception of the world at any resolution.
\newblock \emph{CoRR}, abs/2409.12191, 2024{\natexlab{b}}.

\bibitem[Wang et~al.(2023{\natexlab{c}})Wang, Lv, Yu, Hong, Qi, Wang, Ji, Yang, Zhao, Song, Xu, Xu, Li, Dong, Ding, and Tang]{cogvlm}
Weihan Wang, Qingsong Lv, Wenmeng Yu, Wenyi Hong, Ji~Qi, Yan Wang, Junhui Ji, Zhuoyi Yang, Lei Zhao, Xixuan Song, Jiazheng Xu, Bin Xu, Juanzi Li, Yuxiao Dong, Ming Ding, and Jie Tang.
\newblock Cogvlm: Visual expert for pretrained language models.
\newblock \emph{CoRR}, abs/2311.03079, 2023{\natexlab{c}}.

\bibitem[Wang et~al.(2023{\natexlab{d}})Wang, Chen, Chen, Wu, Zhu, Zeng, Luo, Lu, Zhou, Qiao, and Dai]{visionllm}
Wenhai Wang, Zhe Chen, Xiaokang Chen, Jiannan Wu, Xizhou Zhu, Gang Zeng, Ping Luo, Tong Lu, Jie Zhou, Yu~Qiao, and Jifeng Dai.
\newblock Visionllm: Large language model is also an open-ended decoder for vision-centric tasks.
\newblock In \emph{NeurIPS}, 2023{\natexlab{d}}.

\bibitem[Wang et~al.(2020)Wang, Liu, Shen, Ng, Luo, Jin, Chan, van~den Hengel, and Wang]{estvqa}
Xinyu Wang, Yuliang Liu, Chunhua Shen, Chun~Chet Ng, Canjie Luo, Lianwen Jin, Chee~Seng Chan, Anton van~den Hengel, and Liangwei Wang.
\newblock On the general value of evidence, and bilingual scene-text visual question answering.
\newblock In \emph{CVPR}, 2020.

\bibitem[WebDataset(2024)]{webdataset}
WebDataset.
\newblock Webdataset.
\newblock \url{https://webdataset.github.io/webdataset/}, 2024.

\bibitem[Wei et~al.(2023)Wei, Kong, Chen, Zhao, Ge, Yang, Sun, Han, and Zhang]{vary}
Haoran Wei, Lingyu Kong, Jinyue Chen, Liang Zhao, Zheng Ge, Jinrong Yang, Jianjian Sun, Chunrui Han, and Xiangyu Zhang.
\newblock Vary: Scaling up the vision vocabulary for large vision-language models.
\newblock \emph{CoRR}, abs/2312.06109, 2023.

\bibitem[X.AI(2024)]{rwqa}
X.AI.
\newblock Realworldqa.
\newblock \url{https://huggingface.co/datasets/xai-org/RealworldQA}, 2024.

\bibitem[Xu et~al.(2024)Xu, Yao, Guo, Cui, Ni, Ge, Chua, Liu, Sun, and Huang]{llavauhd}
Ruyi Xu, Yuan Yao, Zonghao Guo, Junbo Cui, Zanlin Ni, Chunjiang Ge, Tat{-}Seng Chua, Zhiyuan Liu, Maosong Sun, and Gao Huang.
\newblock Llava-uhd: an {LMM} perceiving any aspect ratio and high-resolution images.
\newblock \emph{CoRR}, abs/2403.11703, 2024.

\bibitem[Xu et~al.(2022)Xu, Lv, Cui, Wang, Lu, Flor{\^{e}}ncio, Zhang, and Wei]{xfund}
Yiheng Xu, Tengchao Lv, Lei Cui, Guoxin Wang, Yijuan Lu, Dinei A.~F. Flor{\^{e}}ncio, Cha Zhang, and Furu Wei.
\newblock {XFUND:} {A} benchmark dataset for multilingual visually rich form understanding.
\newblock In \emph{ACL}, 2022.

\bibitem[Yan et~al.(2023)Yan, Jiang, Wu, Wang, Luo, Yuan, and Lu]{uninext}
Bin Yan, Yi~Jiang, Jiannan Wu, Dong Wang, Ping Luo, Zehuan Yuan, and Huchuan Lu.
\newblock Universal instance perception as object discovery and retrieval.
\newblock In \emph{CVPR}, 2023.

\bibitem[Yang et~al.(2024)Yang, Yang, Hui, Zheng, Yu, Zhou, Li, Li, Liu, Huang, Dong, Wei, Lin, Tang, Wang, Yang, Tu, Zhang, Ma, Yang, Xu, Zhou, Bai, He, Lin, Dang, Lu, Chen, Yang, Li, Xue, Ni, Zhang, Wang, Peng, Men, Gao, Lin, Wang, Bai, Tan, Zhu, Li, Liu, Ge, Deng, Zhou, Ren, Zhang, Wei, Ren, Liu, Fan, Yao, Zhang, Wan, Chu, Liu, Cui, Zhang, Guo, and Fan]{qwen2}
An~Yang, Baosong Yang, Binyuan Hui, Bo~Zheng, Bowen Yu, Chang Zhou, Chengpeng Li, Chengyuan Li, Dayiheng Liu, Fei Huang, Guanting Dong, Haoran Wei, Huan Lin, Jialong Tang, Jialin Wang, Jian Yang, Jianhong Tu, Jianwei Zhang, Jianxin Ma, Jianxin Yang, Jin Xu, Jingren Zhou, Jinze Bai, Jinzheng He, Junyang Lin, Kai Dang, Keming Lu, Keqin Chen, Kexin Yang, Mei Li, Mingfeng Xue, Na~Ni, Pei Zhang, Peng Wang, Ru~Peng, Rui Men, Ruize Gao, Runji Lin, Shijie Wang, Shuai Bai, Sinan Tan, Tianhang Zhu, Tianhao Li, Tianyu Liu, Wenbin Ge, Xiaodong Deng, Xiaohuan Zhou, Xingzhang Ren, Xinyu Zhang, Xipin Wei, Xuancheng Ren, Xuejing Liu, Yang Fan, Yang Yao, Yichang Zhang, Yu~Wan, Yunfei Chu, Yuqiong Liu, Zeyu Cui, Zhenru Zhang, Zhifang Guo, and Zhihao Fan.
\newblock Qwen2 technical report.
\newblock \emph{CoRR}, abs/2407.10671, 2024.

\bibitem[Yao et~al.(2024)Yao, Yu, Zhang, Wang, Cui, Zhu, Cai, Li, Zhao, He, Chen, Zhou, Zou, Zhang, Hu, Zheng, Zhou, Cai, Han, Zeng, Li, Liu, and Sun]{minicpm-v-2.5}
Yuan Yao, Tianyu Yu, Ao~Zhang, Chongyi Wang, Junbo Cui, Hongji Zhu, Tianchi Cai, Haoyu Li, Weilin Zhao, Zhihui He, Qianyu Chen, Huarong Zhou, Zhensheng Zou, Haoye Zhang, Shengding Hu, Zhi Zheng, Jie Zhou, Jie Cai, Xu~Han, Guoyang Zeng, Dahai Li, Zhiyuan Liu, and Maosong Sun.
\newblock Minicpm-v: {A} {GPT-4V} level {MLLM} on your phone.
\newblock \emph{CoRR}, abs/2408.01800, 2024.

\bibitem[Ye et~al.(2023{\natexlab{a}})Ye, Hu, Xu, Ye, Yan, Dan, Zhao, Xu, Li, Tian, Qi, Zhang, and Huang]{mplugdocowl}
Jiabo Ye, Anwen Hu, Haiyang Xu, Qinghao Ye, Ming Yan, Yuhao Dan, Chenlin Zhao, Guohai Xu, Chenliang Li, Junfeng Tian, Qian Qi, Ji~Zhang, and Fei Huang.
\newblock mplug-docowl: Modularized multimodal large language model for document understanding.
\newblock \emph{CoRR}, abs/2307.02499, 2023{\natexlab{a}}.

\bibitem[Ye et~al.(2023{\natexlab{b}})Ye, Hu, Xu, Ye, Yan, Xu, Li, Tian, Qian, Zhang, Jin, He, Lin, and Huang]{ureader}
Jiabo Ye, Anwen Hu, Haiyang Xu, Qinghao Ye, Ming Yan, Guohai Xu, Chenliang Li, Junfeng Tian, Qi~Qian, Ji~Zhang, Qin Jin, Liang He, Xin Lin, and Fei Huang.
\newblock Ureader: Universal ocr-free visually-situated language understanding with multimodal large language model.
\newblock In \emph{EMNLP}, 2023{\natexlab{b}}.

\bibitem[Ye et~al.(2023{\natexlab{c}})Ye, Xu, Xu, Ye, Yan, Zhou, Wang, Hu, Shi, Shi, Li, Xu, Chen, Tian, Qi, Zhang, and Huang]{mplugowl}
Qinghao Ye, Haiyang Xu, Guohai Xu, Jiabo Ye, Ming Yan, Yiyang Zhou, Junyang Wang, Anwen Hu, Pengcheng Shi, Yaya Shi, Chenliang Li, Yuanhong Xu, Hehong Chen, Junfeng Tian, Qian Qi, Ji~Zhang, and Fei Huang.
\newblock mplug-owl: Modularization empowers large language models with multimodality.
\newblock \emph{CoRR}, abs/2304.14178, 2023{\natexlab{c}}.

\bibitem[Ying et~al.(2024)Ying, Meng, Wang, Li, Lin, Yang, Zhang, Zhang, Lin, Liu, Lei, Lu, Chen, Xu, Zhang, Zhang, Gao, Wang, Qiao, Luo, Zhang, and Shao]{mmtbench}
Kaining Ying, Fanqing Meng, Jin Wang, Zhiqian Li, Han Lin, Yue Yang, Hao Zhang, Wenbo Zhang, Yuqi Lin, Shuo Liu, Jiayi Lei, Quanfeng Lu, Runjian Chen, Peng Xu, Renrui Zhang, Haozhe Zhang, Peng Gao, Yali Wang, Yu~Qiao, Ping Luo, Kaipeng Zhang, and Wenqi Shao.
\newblock Mmt-bench: {A} comprehensive multimodal benchmark for evaluating large vision-language models towards multitask {AGI}.
\newblock In \emph{Proc. ICML}, 2024.

\bibitem[You et~al.(2023)You, Zhang, Gan, Du, Zhang, Wang, Cao, Chang, and Yang]{ferret}
Haoxuan You, Haotian Zhang, Zhe Gan, Xianzhi Du, Bowen Zhang, Zirui Wang, Liangliang Cao, Shih{-}Fu Chang, and Yinfei Yang.
\newblock Ferret: Refer and ground anything anywhere at any granularity.
\newblock \emph{CoRR}, abs/2310.07704, 2023.

\bibitem[Young et~al.(2014)Young, Lai, Hodosh, and Hockenmaier]{flickr}
Peter Young, Alice Lai, Micah Hodosh, and Julia Hockenmaier.
\newblock From image descriptions to visual denotations: New similarity metrics for semantic inference over event descriptions.
\newblock \emph{TACL}, 2, 2014.

\bibitem[Yu et~al.(2023{\natexlab{a}})Yu, Sun, Zhang, Cui, Zhang, Wang, and Liu]{capsfusion}
Qiying Yu, Quan Sun, Xiaosong Zhang, Yufeng Cui, Fan Zhang, Xinlong Wang, and Jingjing Liu.
\newblock Capsfusion: Rethinking image-text data at scale.
\newblock \emph{CoRR}, abs/2310.20550, 2023{\natexlab{a}}.

\bibitem[Yu et~al.(2023{\natexlab{b}})Yu, Yao, Zhang, He, Han, Cui, Hu, Liu, Zheng, Sun, and Chua]{rlhf-v}
Tianyu Yu, Yuan Yao, Haoye Zhang, Taiwen He, Yifeng Han, Ganqu Cui, Jinyi Hu, Zhiyuan Liu, Hai{-}Tao Zheng, Maosong Sun, and Tat{-}Seng Chua.
\newblock {RLHF-V:} towards trustworthy mllms via behavior alignment from fine-grained correctional human feedback.
\newblock \emph{CoRR}, abs/2312.00849, 2023{\natexlab{b}}.

\bibitem[Yu et~al.(2024)Yu, Liao, Wu, Liao, Zheng, and Zeng]{texthawk1}
Ya{-}Qi Yu, Minghui Liao, Jihao Wu, Yongxin Liao, Xiaoyu Zheng, and Wei Zeng.
\newblock {TextHawk}: Exploring efficient fine-grained perception of multimodal large language models.
\newblock \emph{CoRR}, abs/2404.09204, 2024.

\bibitem[Yuan et~al.(2019)Yuan, Zhu, Xu, Li, Mu, and Hu]{ctw}
Tailing Yuan, Zhe Zhu, Kun Xu, Cheng{-}Jun Li, Tai{-}Jiang Mu, and Shi{-}Min Hu.
\newblock A large chinese text dataset in the wild.
\newblock \emph{JCST}, 2019.

\bibitem[Yuan et~al.(2022)Yuan, Liu, Dikubab, Liu, Ji, Wu, and Bai]{hme100k}
Ye~Yuan, Xiao Liu, Wondimu Dikubab, Hui Liu, Zhilong Ji, Zhongqin Wu, and Xiang Bai.
\newblock Syntax-aware network for handwritten mathematical expression recognition.
\newblock In \emph{CVPR}, 2022.

\bibitem[Yue et~al.(2023)Yue, Ni, Zhang, Zheng, Liu, Zhang, Stevens, Jiang, Ren, Sun, Wei, Yu, Yuan, Sun, Yin, Zheng, Yang, Liu, Huang, Sun, Su, and Chen]{mmmu}
Xiang Yue, Yuansheng Ni, Kai Zhang, Tianyu Zheng, Ruoqi Liu, Ge~Zhang, Samuel Stevens, Dongfu Jiang, Weiming Ren, Yuxuan Sun, Cong Wei, Botao Yu, Ruibin Yuan, Renliang Sun, Ming Yin, Boyuan Zheng, Zhenzhu Yang, Yibo Liu, Wenhao Huang, Huan Sun, Yu~Su, and Wenhu Chen.
\newblock {MMMU:} {A} massive multi-discipline multimodal understanding and reasoning benchmark for expert {AGI}.
\newblock \emph{CoRR}, abs/2311.16502, 2023.

\bibitem[Zeng et~al.(2024)Zeng, Xu, Wang, Zhang, Yin, Rojas, Feng, Zhao, Lai, Yu, Wang, Sun, Zhang, Cheng, Gui, Tang, Zhang, Li, Zhao, Wu, Zhong, Liu, Huang, Zhang, Zheng, Lu, Duan, Zhang, Cao, Yang, Tam, Zhao, Liu, Xia, Zhang, Gu, Lv, Liu, Liu, Yang, Song, Zhang, An, Xu, Niu, Yang, Li, Bai, Dong, Qi, Wang, Yang, Du, Hou, and Wang]{glm4}
Aohan Zeng, Bin Xu, Bowen Wang, Chenhui Zhang, Da~Yin, Diego Rojas, Guanyu Feng, Hanlin Zhao, Hanyu Lai, Hao Yu, Hongning Wang, Jiadai Sun, Jiajie Zhang, Jiale Cheng, Jiayi Gui, Jie Tang, Jing Zhang, Juanzi Li, Lei Zhao, Lindong Wu, Lucen Zhong, Mingdao Liu, Minlie Huang, Peng Zhang, Qinkai Zheng, Rui Lu, Shuaiqi Duan, Shudan Zhang, Shulin Cao, Shuxun Yang, Weng~Lam Tam, Wenyi Zhao, Xiao Liu, Xiao Xia, Xiaohan Zhang, Xiaotao Gu, Xin Lv, Xinghan Liu, Xinyi Liu, Xinyue Yang, Xixuan Song, Xunkai Zhang, Yifan An, Yifan Xu, Yilin Niu, Yuantao Yang, Yueyan Li, Yushi Bai, Yuxiao Dong, Zehan Qi, Zhaoyu Wang, Zhen Yang, Zhengxiao Du, Zhenyu Hou, and Zihan Wang.
\newblock Chatglm: {A} family of large language models from {GLM-130B} to {GLM-4} all tools.
\newblock \emph{CoRR}, abs/2406.12793, 2024.

\bibitem[Zhai et~al.(2023)Zhai, Mustafa, Kolesnikov, and Beyer]{siglip}
Xiaohua Zhai, Basil Mustafa, Alexander Kolesnikov, and Lucas Beyer.
\newblock Sigmoid loss for language image pre-training.
\newblock In \emph{ICCV}, 2023.

\bibitem[Zhang et~al.(2023{\natexlab{a}})Zhang, Li, Li, Ren, Zou, Liu, Huang, Gao, Zhang, Li, and Yang]{llava-grounding}
Hao Zhang, Hongyang Li, Feng Li, Tianhe Ren, Xueyan Zou, Shilong Liu, Shijia Huang, Jianfeng Gao, Lei Zhang, Chunyuan Li, and Jianwei Yang.
\newblock Llava-grounding: Grounded visual chat with large multimodal models.
\newblock \emph{CoRR}, abs/2312.02949, 2023{\natexlab{a}}.

\bibitem[Zhang et~al.(2024{\natexlab{a}})Zhang, Gao, Gan, Dufter, Wenzel, Huang, Shah, Du, Zhang, Li, Dodge, You, Yang, Timofeev, Xu, Chen, Fauconnier, Lai, You, Wang, Dehghan, Grasch, and Yang]{mm1.5}
Haotian Zhang, Mingfei Gao, Zhe Gan, Philipp Dufter, Nina Wenzel, Forrest Huang, Dhruti Shah, Xianzhi Du, Bowen Zhang, Yanghao Li, Sam Dodge, Keen You, Zhen Yang, Aleksei Timofeev, Mingze Xu, Hong-You Chen, Jean-Philippe Fauconnier, Zhengfeng Lai, Haoxuan You, Zirui Wang, Afshin Dehghan, Peter Grasch, and Yinfei Yang.
\newblock Mm1.5: Methods, analysis \& insights from multimodal llm fine-tuning.
\newblock \emph{CoRR}, abs/2409.20566, 2024{\natexlab{a}}.

\bibitem[Zhang et~al.(2024{\natexlab{b}})Zhang, You, Dufter, Zhang, Chen, Chen, Fu, Wang, Chang, Gan, and Yang]{ferret2}
Haotian Zhang, Haoxuan You, Philipp Dufter, Bowen Zhang, Chen Chen, Hong{-}You Chen, Tsu{-}Jui Fu, William~Yang Wang, Shih{-}Fu Chang, Zhe Gan, and Yinfei Yang.
\newblock Ferret-v2: An improved baseline for referring and grounding with large language models.
\newblock \emph{CoRR}, abs/2404.07973, 2024{\natexlab{b}}.

\bibitem[Zhang et~al.(2020)Zhang, Liang, and Jin]{hccdoc}
Hesuo Zhang, Lingyu Liang, and Lianwen Jin.
\newblock Scut-hccdoc: {A} new benchmark dataset of handwritten chinese text in unconstrained camera-captured documents.
\newblock \emph{PR}, 2020.

\bibitem[Zhang et~al.(2024{\natexlab{c}})Zhang, Yang, Lai, Xie, and Jin]{dockylin}
Jiaxin Zhang, Wentao Yang, Songxuan Lai, Zecheng Xie, and Lianwen Jin.
\newblock Dockylin: {A} large multimodal model for visual document understanding with efficient visual slimming.
\newblock \emph{CoRR}, abs/2406.19101, 2024{\natexlab{c}}.

\bibitem[Zhang et~al.(2024{\natexlab{d}})Zhang, Yu, Liao, Li, Wu, and Wei]{uihawk}
Jiwen Zhang, Yaqi Yu, Minghui Liao, Wentao Li, Jihao Wu, and Zhongyu Wei.
\newblock {UI-Hawk}: Unleashing the screen stream understanding for gui agents.
\newblock \emph{Preprints}, manuscript/202408.2137, 2024{\natexlab{d}}.

\bibitem[Zhang et~al.(2023{\natexlab{b}})Zhang, Dong, Wang, Cao, Xu, Ouyang, Zhao, Ding, Zhang, Duan, Zhang, Yan, Zhang, Li, Li, Chen, He, Zhang, Qiao, Lin, and Wang]{ixc}
Pan Zhang, Xiaoyi Dong, Bin Wang, Yuhang Cao, Chao Xu, Linke Ouyang, Zhiyuan Zhao, Shuangrui Ding, Songyang Zhang, Haodong Duan, Wenwei Zhang, Hang Yan, Xinyue Zhang, Wei Li, Jingwen Li, Kai Chen, Conghui He, Xingcheng Zhang, Yu~Qiao, Dahua Lin, and Jiaqi Wang.
\newblock Internlm-xcomposer: {A} vision-language large model for advanced text-image comprehension and composition.
\newblock \emph{CoRR}, abs/2309.15112, 2023{\natexlab{b}}.

\bibitem[Zhang et~al.(2019)Zhang, Yang, Bai, Shi, Karatzas, Lu, Jawahar, Zhou, Jiang, Song, Li, Zhou, Wang, Wang, and Liao]{rects}
Rui Zhang, Mingkun Yang, Xiang Bai, Baoguang Shi, Dimosthenis Karatzas, Shijian Lu, C.~V. Jawahar, Yongsheng Zhou, Qianyi Jiang, Qi~Song, Nan Li, Kai Zhou, Lei Wang, Dong Wang, and Minghui Liao.
\newblock {ICDAR} 2019 robust reading challenge on reading chinese text on signboard.
\newblock In \emph{ICDAR}, 2019.

\bibitem[Zhang et~al.(2023{\natexlab{c}})Zhang, Sun, Chen, Xiao, Shao, Zhang, Chen, and Luo]{gpt4roi}
Shilong Zhang, Peize Sun, Shoufa Chen, Min Xiao, Wenqi Shao, Wenwei Zhang, Kai Chen, and Ping Luo.
\newblock Gpt4roi: Instruction tuning large language model on region-of-interest.
\newblock \emph{CoRR}, abs/2307.03601, 2023{\natexlab{c}}.

\bibitem[Zhang et~al.(2023{\natexlab{d}})Zhang, Zhang, Gu, Zhou, Lipka, Yang, and Sun]{llavar}
Yanzhe Zhang, Ruiyi Zhang, Jiuxiang Gu, Yufan Zhou, Nedim Lipka, Diyi Yang, and Tong Sun.
\newblock Llavar: Enhanced visual instruction tuning for text-rich image understanding.
\newblock \emph{CoRR}, abs/2306.17107, 2023{\natexlab{d}}.

\end{thebibliography}


\begin{thebibliography}{78}
\providecommand{\natexlab}[1]{#1}
\providecommand{\url}[1]{\texttt{#1}}
\expandafter\ifx\csname urlstyle\endcsname\relax
  \providecommand{\doi}[1]{doi: #1}\else
  \providecommand{\doi}{doi: \begingroup \urlstyle{rm}\Url}\fi

\bibitem[Anil et~al.(2023)Anil, Borgeaud, Wu, Alayrac, Yu, Soricut, Schalkwyk, Dai, Hauth, Millican, et~al.]{anil2023gemini}
Rohan Anil, Sebastian Borgeaud, Yonghui Wu, Jean-Baptiste Alayrac, Jiahui Yu, Radu Soricut, Johan Schalkwyk, Andrew~M Dai, Anja Hauth, Katie Millican, et~al.
\newblock Gemini: A family of highly capable multimodal models.
\newblock \emph{arXiv preprint arXiv:2312.11805}, 1, 2023.

\bibitem[Austin et~al.(2021)Austin, Johnson, Ho, Tarlow, and Van Den~Berg]{d3pm}
Jacob Austin, Daniel~D Johnson, Jonathan Ho, Daniel Tarlow, and Rianne Van Den~Berg.
\newblock Structured denoising diffusion models in discrete state-spaces.
\newblock \emph{NeurIPS}, pp.\  17981--17993, 2021.

\bibitem[Bai et~al.(2023)Bai, Bai, Yang, Wang, Tan, Wang, Lin, Zhou, and Zhou]{qwen_vl}
Jinze Bai, Shuai Bai, Shusheng Yang, Shijie Wang, Sinan Tan, Peng Wang, Junyang Lin, Chang Zhou, and Jingren Zhou.
\newblock Qwen-vl: {A} frontier large vision-language model with versatile abilities.
\newblock \emph{CoRR}, abs/2308.12966, 2023.

\bibitem[Bai et~al.(2024)Bai, Wang, Xiao, He, Han, Zhang, and Shou]{mllm_hallu}
Zechen Bai, Pichao Wang, Tianjun Xiao, Tong He, Zongbo Han, Zheng Zhang, and Mike~Zheng Shou.
\newblock Hallucination of multimodal large language models: A survey.
\newblock \emph{arXiv preprint arXiv:2404.18930}, 2024.

\bibitem[Brown et~al.(2020)Brown, Mann, Ryder, Subbiah, Kaplan, Dhariwal, Neelakantan, Shyam, Sastry, Askell, Agarwal, Herbert{-}Voss, Krueger, Henighan, Child, Ramesh, Ziegler, Wu, Winter, Hesse, Chen, Sigler, Litwin, Gray, Chess, Clark, Berner, McCandlish, Radford, Sutskever, and Amodei]{gpt3}
Tom~B. Brown, Benjamin Mann, Nick Ryder, Melanie Subbiah, Jared Kaplan, Prafulla Dhariwal, Arvind Neelakantan, Pranav Shyam, Girish Sastry, Amanda Askell, Sandhini Agarwal, Ariel Herbert{-}Voss, Gretchen Krueger, Tom Henighan, Rewon Child, Aditya Ramesh, Daniel~M. Ziegler, Jeffrey Wu, Clemens Winter, Christopher Hesse, Mark Chen, Eric Sigler, Mateusz Litwin, Scott Gray, Benjamin Chess, Jack Clark, Christopher Berner, Sam McCandlish, Alec Radford, Ilya Sutskever, and Dario Amodei.
\newblock Language models are few-shot learners.
\newblock In \emph{NeurIPS}, 2020.

\bibitem[Chang et~al.(2022)Chang, Zhang, Jiang, Liu, and Freeman]{chang2022maskgit}
Huiwen Chang, Han Zhang, Lu~Jiang, Ce~Liu, and William~T Freeman.
\newblock Maskgit: Masked generative image transformer.
\newblock In \emph{CVPR}, pp.\  11315--11325, 2022.

\bibitem[Chang et~al.(2023)Chang, Zhang, Barber, Maschinot, Lezama, Jiang, Yang, Murphy, Freeman, Rubinstein, et~al.]{chang2023muse}
Huiwen Chang, Han Zhang, Jarred Barber, AJ~Maschinot, Jose Lezama, Lu~Jiang, Ming-Hsuan Yang, Kevin Murphy, William~T Freeman, Michael Rubinstein, et~al.
\newblock Muse: Text-to-image generation via masked generative transformers.
\newblock \emph{arXiv preprint arXiv:2301.00704}, 2023.

\bibitem[Changpinyo et~al.(2021)Changpinyo, Sharma, Ding, and Soricut]{cc12m}
Soravit Changpinyo, Piyush Sharma, Nan Ding, and Radu Soricut.
\newblock Conceptual 12m: Pushing web-scale image-text pre-training to recognize long-tail visual concepts.
\newblock In \emph{CVPR}, pp.\  3558--3568, 2021.

\bibitem[Chen et~al.(2024)Chen, Yu, Ge, Yao, Xie, Wang, Kwok, Luo, Lu, and Li]{pixart}
Junsong Chen, Jincheng Yu, Chongjian Ge, Lewei Yao, Enze Xie, Zhongdao Wang, James~T. Kwok, Ping Luo, Huchuan Lu, and Zhenguo Li.
\newblock Pixart-{\(\alpha\)}: Fast training of diffusion transformer for photorealistic text-to-image synthesis.
\newblock In \emph{{ICLR}}. OpenReview.net, 2024.

\bibitem[Chen et~al.(2023)Chen, Li, Dong, Zhang, He, Wang, Zhao, and Lin]{sharegpt4v}
Lin Chen, Jisong Li, Xiaoyi Dong, Pan Zhang, Conghui He, Jiaqi Wang, Feng Zhao, and Dahua Lin.
\newblock Sharegpt4v: Improving large multi-modal models with better captions.
\newblock \emph{arXiv preprint arXiv:2311.12793}, 2023.

\bibitem[Chen et~al.(2020)Chen, Radford, Child, Wu, Jun, Luan, and Sutskever]{chen2020generative}
Mark Chen, Alec Radford, Rewon Child, Jeffrey Wu, Heewoo Jun, David Luan, and Ilya Sutskever.
\newblock Generative pretraining from pixels.
\newblock In \emph{ICML}, pp.\  1691--1703, 2020.

\bibitem[Chowdhery et~al.(2023)Chowdhery, Narang, Devlin, Bosma, Mishra, Roberts, Barham, Chung, Sutton, Gehrmann, Schuh, Shi, Tsvyashchenko, Maynez, Rao, Barnes, Tay, Shazeer, Prabhakaran, Reif, Du, Hutchinson, Pope, Bradbury, Austin, Isard, Gur{-}Ari, Yin, Duke, Levskaya, Ghemawat, Dev, Michalewski, Garcia, Misra, Robinson, Fedus, Zhou, Ippolito, Luan, Lim, Zoph, Spiridonov, Sepassi, Dohan, Agrawal, Omernick, Dai, Pillai, Pellat, Lewkowycz, Moreira, Child, Polozov, Lee, Zhou, Wang, Saeta, Diaz, Firat, Catasta, Wei, Meier{-}Hellstern, Eck, Dean, Petrov, and Fiedel]{PALM}
Aakanksha Chowdhery, Sharan Narang, Jacob Devlin, Maarten Bosma, Gaurav Mishra, Adam Roberts, Paul Barham, Hyung~Won Chung, Charles Sutton, Sebastian Gehrmann, Parker Schuh, Kensen Shi, Sasha Tsvyashchenko, Joshua Maynez, Abhishek Rao, Parker Barnes, Yi~Tay, Noam Shazeer, Vinodkumar Prabhakaran, Emily Reif, Nan Du, Ben Hutchinson, Reiner Pope, James Bradbury, Jacob Austin, Michael Isard, Guy Gur{-}Ari, Pengcheng Yin, Toju Duke, Anselm Levskaya, Sanjay Ghemawat, Sunipa Dev, Henryk Michalewski, Xavier Garcia, Vedant Misra, Kevin Robinson, Liam Fedus, Denny Zhou, Daphne Ippolito, David Luan, Hyeontaek Lim, Barret Zoph, Alexander Spiridonov, Ryan Sepassi, David Dohan, Shivani Agrawal, Mark Omernick, Andrew~M. Dai, Thanumalayan~Sankaranarayana Pillai, Marie Pellat, Aitor Lewkowycz, Erica Moreira, Rewon Child, Oleksandr Polozov, Katherine Lee, Zongwei Zhou, Xuezhi Wang, Brennan Saeta, Mark Diaz, Orhan Firat, Michele Catasta, Jason Wei, Kathy Meier{-}Hellstern, Douglas Eck, Jeff Dean, Slav Petrov, and Noah Fiedel.
\newblock Palm: Scaling language modeling with pathways.
\newblock \emph{J. Mach. Learn. Res.}, 24:\penalty0 240:1--240:113, 2023.

\bibitem[Dai et~al.(2023)Dai, Li, Li, Tiong, Zhao, Wang, Li, Fung, and Hoi]{instructblip}
Wenliang Dai, Junnan Li, Dongxu Li, Anthony Meng~Huat Tiong, Junqi Zhao, Weisheng Wang, Boyang Li, Pascale Fung, and Steven Hoi.
\newblock Instructblip: Towards general-purpose vision-language models with instruction tuning, 2023.

\bibitem[Dehghani et~al.(2023)Dehghani, Djolonga, Mustafa, Padlewski, Heek, Gilmer, Steiner, Caron, Geirhos, Alabdulmohsin, Jenatton, Beyer, Tschannen, Arnab, Wang, Ruiz, Minderer, Puigcerver, Evci, Kumar, van Steenkiste, Elsayed, Mahendran, Yu, Oliver, Huot, Bastings, Collier, Gritsenko, Birodkar, Vasconcelos, Tay, Mensink, Kolesnikov, Pavetic, Tran, Kipf, Lucic, Zhai, Keysers, Harmsen, and Houlsby]{qknorm}
Mostafa Dehghani, Josip Djolonga, Basil Mustafa, Piotr Padlewski, Jonathan Heek, Justin Gilmer, Andreas~Peter Steiner, Mathilde Caron, Robert Geirhos, Ibrahim Alabdulmohsin, Rodolphe Jenatton, Lucas Beyer, Michael Tschannen, Anurag Arnab, Xiao Wang, Carlos~Riquelme Ruiz, Matthias Minderer, Joan Puigcerver, Utku Evci, Manoj Kumar, Sjoerd van Steenkiste, Gamaleldin~Fathy Elsayed, Aravindh Mahendran, Fisher Yu, Avital Oliver, Fantine Huot, Jasmijn Bastings, Mark Collier, Alexey~A. Gritsenko, Vighnesh Birodkar, Cristina~Nader Vasconcelos, Yi~Tay, Thomas Mensink, Alexander Kolesnikov, Filip Pavetic, Dustin Tran, Thomas Kipf, Mario Lucic, Xiaohua Zhai, Daniel Keysers, Jeremiah~J. Harmsen, and Neil Houlsby.
\newblock Scaling vision transformers to 22 billion parameters.
\newblock In \emph{ICML}, pp.\  7480--7512, 2023.

\bibitem[Deng et~al.(2009)Deng, Dong, Socher, Li, Li, and Fei-Fei]{imagenet}
Jia Deng, Wei Dong, Richard Socher, Li-Jia Li, Kai Li, and Li~Fei-Fei.
\newblock Imagenet: A large-scale hierarchical image database.
\newblock In \emph{CVPR}, pp.\  248--255, 2009.

\bibitem[Esser et~al.(2021{\natexlab{a}})Esser, Rombach, and Ommer]{taming}
Patrick Esser, Robin Rombach, and Bjorn Ommer.
\newblock Taming transformers for high-resolution image synthesis.
\newblock In \emph{CVPR}, pp.\  12873--12883, 2021{\natexlab{a}}.

\bibitem[Esser et~al.(2021{\natexlab{b}})Esser, Rombach, and Ommer]{vqgan}
Patrick Esser, Robin Rombach, and Bjorn Ommer.
\newblock Taming transformers for high-resolution image synthesis.
\newblock In \emph{CVPR}, pp.\  12873--12883, 2021{\natexlab{b}}.

\bibitem[Esser et~al.(2024)Esser, Kulal, Blattmann, Entezari, M{\"u}ller, Saini, Levi, Lorenz, Sauer, Boesel, et~al.]{sd3}
Patrick Esser, Sumith Kulal, Andreas Blattmann, Rahim Entezari, Jonas M{\"u}ller, Harry Saini, Yam Levi, Dominik Lorenz, Axel Sauer, Frederic Boesel, et~al.
\newblock Scaling rectified flow transformers for high-resolution image synthesis.
\newblock In \emph{ICML}, 2024.

\bibitem[Ge et~al.(2024)Ge, Zhao, Zhu, Ge, Yi, Song, Li, Ding, and Shan]{seed-x}
Yuying Ge, Sijie Zhao, Jinguo Zhu, Yixiao Ge, Kun Yi, Lin Song, Chen Li, Xiaohan Ding, and Ying Shan.
\newblock Seed-x: Multimodal models with unified multi-granularity comprehension and generation.
\newblock \emph{arXiv preprint arXiv:2404.14396}, 2024.

\bibitem[Ghazvininejad et~al.(2019)Ghazvininejad, Levy, Liu, and Zettlemoyer]{maskpredict}
Marjan Ghazvininejad, Omer Levy, Yinhan Liu, and Luke Zettlemoyer.
\newblock Mask-predict: Parallel decoding of conditional masked language models.
\newblock In \emph{EMNLP}, pp.\  6111--6120, 2019.

\bibitem[Ghosh et~al.(2023)Ghosh, Hajishirzi, and Schmidt]{geneval}
Dhruba Ghosh, Hannaneh Hajishirzi, and Ludwig Schmidt.
\newblock Geneval: An object-focused framework for evaluating text-to-image alignment.
\newblock In \emph{NeurIPS}, 2023.

\bibitem[Goodfellow et~al.(2014)Goodfellow, Pouget-Abadie, Mirza, Xu, Warde-Farley, Ozair, Courville, and Bengio]{gan}
Ian Goodfellow, Jean Pouget-Abadie, Mehdi Mirza, Bing Xu, David Warde-Farley, Sherjil Ozair, Aaron Courville, and Yoshua Bengio.
\newblock Generative adversarial nets.
\newblock \emph{NeurIPS}, 2014.

\bibitem[Gu et~al.(2022)Gu, Chen, Bao, Wen, Zhang, Chen, Yuan, and Guo]{vqdiffusion}
Shuyang Gu, Dong Chen, Jianmin Bao, Fang Wen, Bo~Zhang, Dongdong Chen, Lu~Yuan, and Baining Guo.
\newblock Vector quantized diffusion model for text-to-image synthesis.
\newblock In \emph{CVPR}, pp.\  10696--10706, 2022.

\bibitem[Gu et~al.(2024)Gu, Wang, Ge, Shan, and Shou]{gu2024rethinking}
Yuchao Gu, Xintao Wang, Yixiao Ge, Ying Shan, and Mike~Zheng Shou.
\newblock Rethinking the objectives of vector-quantized tokenizers for image synthesis.
\newblock In \emph{Proceedings of the IEEE/CVF Conference on Computer Vision and Pattern Recognition}, pp.\  7631--7640, 2024.

\bibitem[Ho \& Salimans(2022)Ho and Salimans]{ho2022classifier}
Jonathan Ho and Tim Salimans.
\newblock Classifier-free diffusion guidance.
\newblock \emph{arXiv preprint arXiv:2207.12598}, 2022.

\bibitem[Ho et~al.(2020)Ho, Jain, and Abbeel]{ho2020denoising}
Jonathan Ho, Ajay Jain, and Pieter Abbeel.
\newblock Denoising diffusion probabilistic models.
\newblock \emph{NeurIPS}, pp.\  6840--6851, 2020.

\bibitem[Ho et~al.(2022)Ho, Salimans, Gritsenko, Chan, Norouzi, and Fleet]{ho2022video}
Jonathan Ho, Tim Salimans, Alexey Gritsenko, William Chan, Mohammad Norouzi, and David~J Fleet.
\newblock Video diffusion models.
\newblock \emph{NeurIPS}, 2022.

\bibitem[Hoogeboom et~al.(2022)Hoogeboom, Gritsenko, Bastings, Poole, van~den Berg, and Salimans]{ARDM}
Emiel Hoogeboom, Alexey~A. Gritsenko, Jasmijn Bastings, Ben Poole, Rianne van~den Berg, and Tim Salimans.
\newblock Autoregressive diffusion models.
\newblock In \emph{{ICLR}}. OpenReview.net, 2022.

\bibitem[Kang et~al.(2023)Kang, Zhu, Zhang, Park, Shechtman, Paris, and Park]{gigagan}
Minguk Kang, Jun{-}Yan Zhu, Richard Zhang, Jaesik Park, Eli Shechtman, Sylvain Paris, and Taesung Park.
\newblock Scaling up gans for text-to-image synthesis.
\newblock In \emph{{CVPR}}, pp.\  10124--10134. {IEEE}, 2023.

\bibitem[Kingma \& Welling(2013)Kingma and Welling]{vae}
Diederik~P Kingma and Max Welling.
\newblock Auto-encoding variational bayes.
\newblock \emph{arXiv preprint arXiv:1312.6114}, 2013.

\bibitem[Kirillov et~al.(2023)Kirillov, Mintun, Ravi, Mao, Rolland, Gustafson, Xiao, Whitehead, Berg, Lo, et~al.]{sa1b}
Alexander Kirillov, Eric Mintun, Nikhila Ravi, Hanzi Mao, Chloe Rolland, Laura Gustafson, Tete Xiao, Spencer Whitehead, Alexander~C Berg, Wan-Yen Lo, et~al.
\newblock Segment anything.
\newblock In \emph{ICCV}, pp.\  4015--4026, 2023.

\bibitem[Kondratyuk et~al.(2023)Kondratyuk, Yu, Gu, Lezama, Huang, Hornung, Adam, Akbari, Alon, Birodkar, et~al.]{kondratyuk2023videopoet}
Dan Kondratyuk, Lijun Yu, Xiuye Gu, Jos{\'e} Lezama, Jonathan Huang, Rachel Hornung, Hartwig Adam, Hassan Akbari, Yair Alon, Vighnesh Birodkar, et~al.
\newblock Videopoet: A large language model for zero-shot video generation.
\newblock \emph{arXiv preprint arXiv:2312.14125}, 2023.

\bibitem[Li et~al.(2024)Li, Gan, Yang, Yang, Li, Wang, Gao, et~al.]{li2024multimodal}
Chunyuan Li, Zhe Gan, Zhengyuan Yang, Jianwei Yang, Linjie Li, Lijuan Wang, Jianfeng Gao, et~al.
\newblock Multimodal foundation models: From specialists to general-purpose assistants.
\newblock \emph{Foundations and Trends{\textregistered} in Computer Graphics and Vision}, 16\penalty0 (1-2):\penalty0 1--214, 2024.

\bibitem[Li et~al.(2023)Li, Bubeck, Eldan, Del~Giorno, Gunasekar, and Lee]{phi1.5}
Yuanzhi Li, S{\'e}bastien Bubeck, Ronen Eldan, Allie Del~Giorno, Suriya Gunasekar, and Yin~Tat Lee.
\newblock Textbooks are all you need ii: phi-1.5 technical report.
\newblock \emph{arXiv preprint arXiv:2309.05463}, 2023.

\bibitem[Liu et~al.(2024{\natexlab{a}})Liu, Yan, Zaharia, and Abbeel]{lwm}
Hao Liu, Wilson Yan, Matei Zaharia, and Pieter Abbeel.
\newblock World model on million-length video and language with ringattention.
\newblock \emph{arXiv preprint}, 2024{\natexlab{a}}.

\bibitem[Liu et~al.(2024{\natexlab{b}})Liu, Li, Li, and Lee]{llava1.5}
Haotian Liu, Chunyuan Li, Yuheng Li, and Yong~Jae Lee.
\newblock Improved baselines with visual instruction tuning.
\newblock In \emph{CVPR}, pp.\  26296--26306, 2024{\natexlab{b}}.

\bibitem[Liu et~al.(2024{\natexlab{c}})Liu, Li, Wu, and Lee]{llava}
Haotian Liu, Chunyuan Li, Qingyang Wu, and Yong~Jae Lee.
\newblock Visual instruction tuning.
\newblock \emph{NeurIPS}, 36, 2024{\natexlab{c}}.

\bibitem[McKinzie et~al.(2024)McKinzie, Gan, Fauconnier, Dodge, Zhang, Dufter, Shah, Du, Peng, Weers, et~al.]{mckinzie2024mm1}
Brandon McKinzie, Zhe Gan, Jean-Philippe Fauconnier, Sam Dodge, Bowen Zhang, Philipp Dufter, Dhruti Shah, Xianzhi Du, Futang Peng, Floris Weers, et~al.
\newblock Mm1: Methods, analysis \& insights from multimodal llm pre-training.
\newblock \emph{arXiv preprint arXiv:2403.09611}, 2024.

\bibitem[Murphy(2023)]{pml2Book}
Kevin~P. Murphy.
\newblock \emph{Probabilistic Machine Learning: Advanced Topics}.
\newblock MIT Press, 2023.
\newblock URL \url{http://probml.github.io/book2}.

\bibitem[Nichol et~al.(2021)Nichol, Dhariwal, Ramesh, Shyam, Mishkin, McGrew, Sutskever, and Chen]{nichol2021glide}
Alex Nichol, Prafulla Dhariwal, Aditya Ramesh, Pranav Shyam, Pamela Mishkin, Bob McGrew, Ilya Sutskever, and Mark Chen.
\newblock Glide: Towards photorealistic image generation and editing with text-guided diffusion models.
\newblock \emph{arXiv preprint arXiv:2112.10741}, 2021.

\bibitem[Parmar et~al.(2018)Parmar, Vaswani, Uszkoreit, Kaiser, Shazeer, Ku, and Tran]{parmar2018image}
Niki Parmar, Ashish Vaswani, Jakob Uszkoreit, Lukasz Kaiser, Noam Shazeer, Alexander Ku, and Dustin Tran.
\newblock Image transformer.
\newblock In \emph{ICML}, pp.\  4055--4064, 2018.

\bibitem[Peebles \& Xie(2023)Peebles and Xie]{peebles2023scalable}
William Peebles and Saining Xie.
\newblock Scalable diffusion models with transformers.
\newblock In \emph{ICCV}, pp.\  4195--4205, 2023.

\bibitem[Penedo et~al.(2023)Penedo, Malartic, Hesslow, Cojocaru, Alobeidli, Cappelli, Pannier, Almazrouei, and Launay]{refinedweb}
Guilherme Penedo, Quentin Malartic, Daniel Hesslow, Ruxandra Cojocaru, Hamza Alobeidli, Alessandro Cappelli, Baptiste Pannier, Ebtesam Almazrouei, and Julien Launay.
\newblock The refinedweb dataset for falcon {LLM:} outperforming curated corpora with web data only.
\newblock In \emph{NeurIPS}, 2023.

\bibitem[Podell et~al.(2023)Podell, English, Lacey, Blattmann, Dockhorn, M{\"u}ller, Penna, and Rombach]{sdxl}
Dustin Podell, Zion English, Kyle Lacey, Andreas Blattmann, Tim Dockhorn, Jonas M{\"u}ller, Joe Penna, and Robin Rombach.
\newblock Sdxl: Improving latent diffusion models for high-resolution image synthesis.
\newblock \emph{arXiv preprint arXiv:2307.01952}, 2023.

\bibitem[Radford et~al.(2018)Radford, Narasimhan, Salimans, Sutskever, et~al.]{gpt1}
Alec Radford, Karthik Narasimhan, Tim Salimans, Ilya Sutskever, et~al.
\newblock Improving language understanding by generative pre-training.
\newblock 2018.

\bibitem[Radford et~al.(2021)Radford, Kim, Hallacy, Ramesh, Goh, Agarwal, Sastry, Askell, Mishkin, Clark, Krueger, and Sutskever]{clip}
Alec Radford, Jong~Wook Kim, Chris Hallacy, Aditya Ramesh, Gabriel Goh, Sandhini Agarwal, Girish Sastry, Amanda Askell, Pamela Mishkin, Jack Clark, Gretchen Krueger, and Ilya Sutskever.
\newblock Learning transferable visual models from natural language supervision.
\newblock In \emph{{ICML}}, pp.\  8748--8763, 2021.

\bibitem[Raffel et~al.(2020)Raffel, Shazeer, Roberts, Lee, Narang, Matena, Zhou, Li, and Liu]{raffel2020exploring}
Colin Raffel, Noam Shazeer, Adam Roberts, Katherine Lee, Sharan Narang, Michael Matena, Yanqi Zhou, Wei Li, and Peter~J Liu.
\newblock Exploring the limits of transfer learning with a unified text-to-text transformer.
\newblock \emph{Journal of machine learning research}, 21\penalty0 (140):\penalty0 1--67, 2020.

\bibitem[Ramesh et~al.(2021)Ramesh, Pavlov, Goh, Gray, Voss, Radford, Chen, and Sutskever]{dalle}
Aditya Ramesh, Mikhail Pavlov, Gabriel Goh, Scott Gray, Chelsea Voss, Alec Radford, Mark Chen, and Ilya Sutskever.
\newblock Zero-shot text-to-image generation.
\newblock In \emph{ICML}, pp.\  8821--8831. Pmlr, 2021.

\bibitem[Ramesh et~al.(2022{\natexlab{a}})Ramesh, Dhariwal, Nichol, Chu, and Chen]{dalle2}
Aditya Ramesh, Prafulla Dhariwal, Alex Nichol, Casey Chu, and Mark Chen.
\newblock Hierarchical text-conditional image generation with {CLIP} latents.
\newblock \emph{CoRR}, abs/2204.06125, 2022{\natexlab{a}}.

\bibitem[Ramesh et~al.(2022{\natexlab{b}})Ramesh, Dhariwal, Nichol, Chu, and Chen]{ramesh2022hierarchical}
Aditya Ramesh, Prafulla Dhariwal, Alex Nichol, Casey Chu, and Mark Chen.
\newblock Hierarchical text-conditional image generation with clip latents.
\newblock \emph{arXiv preprint arXiv:2204.06125}, 1\penalty0 (2):\penalty0 3, 2022{\natexlab{b}}.

\bibitem[Ravuri \& Vinyals(2019)Ravuri and Vinyals]{ravuri2019classification}
Suman Ravuri and Oriol Vinyals.
\newblock Classification accuracy score for conditional generative models.
\newblock \emph{NeurIPS}, 32, 2019.

\bibitem[Rombach et~al.(2022)Rombach, Blattmann, Lorenz, Esser, and Ommer]{rombach2022high}
Robin Rombach, Andreas Blattmann, Dominik Lorenz, Patrick Esser, and Bj{\"o}rn Ommer.
\newblock High-resolution image synthesis with latent diffusion models.
\newblock In \emph{CVPR}, pp.\  10684--10695, 2022.

\bibitem[Saharia et~al.(2022)Saharia, Chan, Saxena, Li, Whang, Denton, Ghasemipour, Gontijo~Lopes, Karagol~Ayan, Salimans, et~al.]{imagen}
Chitwan Saharia, William Chan, Saurabh Saxena, Lala Li, Jay Whang, Emily~L Denton, Kamyar Ghasemipour, Raphael Gontijo~Lopes, Burcu Karagol~Ayan, Tim Salimans, et~al.
\newblock Photorealistic text-to-image diffusion models with deep language understanding.
\newblock \emph{NeurIPS}, 35:\penalty0 36479--36494, 2022.

\bibitem[Sohl-Dickstein et~al.(2015)Sohl-Dickstein, Weiss, Maheswaranathan, and Ganguli]{sohl2015deep}
Jascha Sohl-Dickstein, Eric Weiss, Niru Maheswaranathan, and Surya Ganguli.
\newblock Deep unsupervised learning using nonequilibrium thermodynamics.
\newblock In \emph{ICML}, pp.\  2256--2265, 2015.

\bibitem[Sou\v{c}ek et~al.(2024)Sou\v{c}ek, Damen, Wray, Laptev, and Sivic]{genhowto}
Tom\'a\v{s} Sou\v{c}ek, Dima Damen, Michael Wray, Ivan Laptev, and Josef Sivic.
\newblock Genhowto: Learning to generate actions and state transformations from instructional videos.
\newblock In \emph{CVPR}, pp.\  6561--6571, 2024.

\bibitem[Sun et~al.(2023{\natexlab{a}})Sun, Pan, Ge, Li, Duan, Wu, Zhang, Zhou, Qin, Wang, Dai, Qiao, Wang, and Li]{journeydb}
Keqiang Sun, Junting Pan, Yuying Ge, Hao Li, Haodong Duan, Xiaoshi Wu, Renrui Zhang, Aojun Zhou, Zipeng Qin, Yi~Wang, Jifeng Dai, Yu~Qiao, Limin Wang, and Hongsheng Li.
\newblock Journeydb: {A} benchmark for generative image understanding.
\newblock In \emph{NeurIPS}, 2023{\natexlab{a}}.

\bibitem[Sun et~al.(2024)Sun, Jiang, Chen, Zhang, Peng, Luo, and Yuan]{llamagen}
Peize Sun, Yi~Jiang, Shoufa Chen, Shilong Zhang, Bingyue Peng, Ping Luo, and Zehuan Yuan.
\newblock Autoregressive model beats diffusion: Llama for scalable image generation.
\newblock \emph{arXiv preprint arXiv:2406.06525}, 2024.

\bibitem[Sun et~al.(2023{\natexlab{b}})Sun, Cui, Zhang, Zhang, Yu, Luo, Wang, Rao, Liu, Huang, and Wang]{DBLP:journals/corr/abs-2312-13286}
Quan Sun, Yufeng Cui, Xiaosong Zhang, Fan Zhang, Qiying Yu, Zhengxiong Luo, Yueze Wang, Yongming Rao, Jingjing Liu, Tiejun Huang, and Xinlong Wang.
\newblock Generative multimodal models are in-context learners.
\newblock \emph{CoRR}, abs/2312.13286, 2023{\natexlab{b}}.

\bibitem[Sun et~al.(2023{\natexlab{c}})Sun, Yu, Cui, Zhang, Zhang, Wang, Gao, Liu, Huang, and Wang]{DBLP:journals/corr/abs-2307-05222}
Quan Sun, Qiying Yu, Yufeng Cui, Fan Zhang, Xiaosong Zhang, Yueze Wang, Hongcheng Gao, Jingjing Liu, Tiejun Huang, and Xinlong Wang.
\newblock Generative pretraining in multimodality.
\newblock \emph{CoRR}, abs/2307.05222, 2023{\natexlab{c}}.

\bibitem[Sun et~al.(2023{\natexlab{d}})Sun, Yu, Cui, Zhang, Zhang, Wang, Gao, Liu, Huang, and Wang]{sun2023emu}
Quan Sun, Qiying Yu, Yufeng Cui, Fan Zhang, Xiaosong Zhang, Yueze Wang, Hongcheng Gao, Jingjing Liu, Tiejun Huang, and Xinlong Wang.
\newblock Emu: Generative pretraining in multimodality.
\newblock In \emph{ICLR}, 2023{\natexlab{d}}.

\bibitem[Tang et~al.(2024)Tang, Yang, Zhu, Zeng, and Bansal]{CoDI}
Zineng Tang, Ziyi Yang, Chenguang Zhu, Michael Zeng, and Mohit Bansal.
\newblock Any-to-any generation via composable diffusion.
\newblock \emph{NeurIPS}, 36, 2024.

\bibitem[Team(2024)]{team2024chameleon}
Chameleon Team.
\newblock Chameleon: Mixed-modal early-fusion foundation models.
\newblock \emph{arXiv preprint arXiv:2405.09818}, 2024.

\bibitem[Tong et~al.(2024)Tong, Brown, Wu, Woo, Middepogu, Akula, Yang, Yang, Iyer, Pan, et~al.]{tong2024cambrian}
Shengbang Tong, Ellis Brown, Penghao Wu, Sanghyun Woo, Manoj Middepogu, Sai~Charitha Akula, Jihan Yang, Shusheng Yang, Adithya Iyer, Xichen Pan, et~al.
\newblock Cambrian-1: A fully open, vision-centric exploration of multimodal llms.
\newblock \emph{arXiv preprint arXiv:2406.16860}, 2024.

\bibitem[Touvron et~al.(2023)Touvron, Lavril, Izacard, Martinet, Lachaux, Lacroix, Rozi{\`{e}}re, Goyal, Hambro, Azhar, Rodriguez, Joulin, Grave, and Lample]{llama}
Hugo Touvron, Thibaut Lavril, Gautier Izacard, Xavier Martinet, Marie{-}Anne Lachaux, Timoth{\'{e}}e Lacroix, Baptiste Rozi{\`{e}}re, Naman Goyal, Eric Hambro, Faisal Azhar, Aur{\'{e}}lien Rodriguez, Armand Joulin, Edouard Grave, and Guillaume Lample.
\newblock Llama: Open and efficient foundation language models.
\newblock \emph{CoRR}, abs/2302.13971, 2023.

\bibitem[Vaswani et~al.(2017{\natexlab{a}})Vaswani, Shazeer, Parmar, Uszkoreit, Jones, Gomez, Kaiser, and Polosukhin]{attention}
Ashish Vaswani, Noam Shazeer, Niki Parmar, Jakob Uszkoreit, Llion Jones, Aidan~N Gomez, {\L}ukasz Kaiser, and Illia Polosukhin.
\newblock Attention is all you need.
\newblock \emph{NeurIPS}, 30, 2017{\natexlab{a}}.

\bibitem[Vaswani et~al.(2017{\natexlab{b}})Vaswani, Shazeer, Parmar, Uszkoreit, Jones, Gomez, Kaiser, and Polosukhin]{transformer}
Ashish Vaswani, Noam Shazeer, Niki Parmar, Jakob Uszkoreit, Llion Jones, Aidan~N Gomez, {\L}ukasz Kaiser, and Illia Polosukhin.
\newblock Attention is all you need.
\newblock \emph{NeurIPS}, 2017{\natexlab{b}}.

\bibitem[Wang \& Cho(2019)Wang and Cho]{wang2019bert}
Alex Wang and Kyunghyun Cho.
\newblock Bert has a mouth, and it must speak: Bert as a markov random field language model.
\newblock \emph{arXiv preprint arXiv:1902.04094}, 2019.

\bibitem[Wortsman et~al.(2023)Wortsman, Liu, Xiao, Everett, Alemi, Adlam, Co-Reyes, Gur, Kumar, Novak, et~al.]{qknorm2}
Mitchell Wortsman, Peter~J Liu, Lechao Xiao, Katie Everett, Alex Alemi, Ben Adlam, John~D Co-Reyes, Izzeddin Gur, Abhishek Kumar, Roman Novak, et~al.
\newblock Small-scale proxies for large-scale transformer training instabilities.
\newblock \emph{arXiv preprint arXiv:2309.14322}, 2023.

\bibitem[Wu et~al.(2023{\natexlab{a}})Wu, Ge, Wang, Lei, Gu, Shi, Hsu, Shan, Qie, and Shou]{wu2023tune}
Jay~Zhangjie Wu, Yixiao Ge, Xintao Wang, Stan~Weixian Lei, Yuchao Gu, Yufei Shi, Wynne Hsu, Ying Shan, Xiaohu Qie, and Mike~Zheng Shou.
\newblock Tune-a-video: One-shot tuning of image diffusion models for text-to-video generation.
\newblock In \emph{ICCV}, 2023{\natexlab{a}}.

\bibitem[Wu et~al.(2023{\natexlab{b}})Wu, Fei, Qu, Ji, and Chua]{wu2023next}
Shengqiong Wu, Hao Fei, Leigang Qu, Wei Ji, and Tat-Seng Chua.
\newblock Next-gpt: Any-to-any multimodal llm.
\newblock \emph{arXiv preprint arXiv:2309.05519}, 2023{\natexlab{b}}.

\bibitem[Xie et~al.(2023)Xie, Li, Huang, Liu, Zhang, Zheng, and Shou]{Xie_2023_ICCV}
Jinheng Xie, Yuexiang Li, Yawen Huang, Haozhe Liu, Wentian Zhang, Yefeng Zheng, and Mike~Zheng Shou.
\newblock Boxdiff: Text-to-image synthesis with training-free box-constrained diffusion.
\newblock In \emph{ICCV}, pp.\  7452--7461, 2023.

\bibitem[Xue et~al.(2024)Xue, Song, Guo, Liu, Zong, Liu, and Luo]{raphael}
Zeyue Xue, Guanglu Song, Qiushan Guo, Boxiao Liu, Zhuofan Zong, Yu~Liu, and Ping Luo.
\newblock Raphael: Text-to-image generation via large mixture of diffusion paths.
\newblock \emph{NeurIPS}, 36, 2024.

\bibitem[Ye et~al.(2024{\natexlab{a}})Ye, Huang, Lu, Yu, Ping, Tao, Kautz, Han, Xu, Molchanov, et~al.]{x-vila}
Hanrong Ye, De-An Huang, Yao Lu, Zhiding Yu, Wei Ping, Andrew Tao, Jan Kautz, Song Han, Dan Xu, Pavlo Molchanov, et~al.
\newblock X-vila: Cross-modality alignment for large language model.
\newblock \emph{arXiv preprint arXiv:2405.19335}, 2024{\natexlab{a}}.

\bibitem[Ye et~al.(2024{\natexlab{b}})Ye, Xu, Ye, Yan, Hu, Liu, Qian, Zhang, and Huang]{ye2024mplug}
Qinghao Ye, Haiyang Xu, Jiabo Ye, Ming Yan, Anwen Hu, Haowei Liu, Qi~Qian, Ji~Zhang, and Fei Huang.
\newblock mplug-owl2: Revolutionizing multi-modal large language model with modality collaboration.
\newblock In \emph{CVPR}, pp.\  13040--13051, 2024{\natexlab{b}}.

\bibitem[Yin et~al.(2023)Yin, Fu, Zhao, Li, Sun, Xu, and Chen]{mllm_survey}
Shukang Yin, Chaoyou Fu, Sirui Zhao, Ke~Li, Xing Sun, Tong Xu, and Enhong Chen.
\newblock A survey on multimodal large language models.
\newblock \emph{arXiv preprint arXiv:2306.13549}, 2023.

\bibitem[Yu et~al.(2023)Yu, Lezama, Gundavarapu, Versari, Sohn, Minnen, Cheng, Gupta, Gu, Hauptmann, et~al.]{magvitv2}
Lijun Yu, Jos{\'e} Lezama, Nitesh~B Gundavarapu, Luca Versari, Kihyuk Sohn, David Minnen, Yong Cheng, Agrim Gupta, Xiuye Gu, Alexander~G Hauptmann, et~al.
\newblock Language model beats diffusion--tokenizer is key to visual generation.
\newblock \emph{arXiv preprint arXiv:2310.05737}, 2023.

\bibitem[Zhu et~al.(2023{\natexlab{a}})Zhu, Chen, Shen, Li, and Elhoseiny]{minigpt4}
Deyao Zhu, Jun Chen, Xiaoqian Shen, Xiang Li, and Mohamed Elhoseiny.
\newblock Minigpt-4: Enhancing vision-language understanding with advanced large language models.
\newblock \emph{CoRR}, abs/2304.10592, 2023{\natexlab{a}}.

\bibitem[Zhu et~al.(2023{\natexlab{b}})Zhu, Ding, Ge, Ge, Zhao, Zhao, Wang, and Shan]{vlgpt}
Jinguo Zhu, Xiaohan Ding, Yixiao Ge, Yuying Ge, Sijie Zhao, Hengshuang Zhao, Xiaohua Wang, and Ying Shan.
\newblock {VL-GPT:} {A} generative pre-trained transformer for vision and language understanding and generation.
\newblock \emph{CoRR}, abs/2312.09251, 2023{\natexlab{b}}.

\end{thebibliography}
